\documentclass{article}

\usepackage[preprint]{neurips_2025}

\usepackage[utf8]{inputenc} 
\usepackage[T1]{fontenc}    
\usepackage{hyperref}       
\usepackage{url}            
\usepackage{booktabs}       
\usepackage{amsfonts}       
\usepackage{nicefrac}       
\usepackage{microtype}      
\usepackage{xcolor}         
\usepackage{amssymb}
\usepackage{graphicx}
\usepackage{amsthm}
\usepackage{cleveref}
\usepackage{wrapfig}
\usepackage{placeins}

\newtheorem{remark}{Remark}[section]

\newcommand{\D}{\mathcal{D}} 

\newcommand{\Model}{f}
\newcommand{\Loss}{\mathcal{L}} 

\newcommand{\ys}{\mathbf{y}} 
\newcommand{\xs}{\mathbf{x}}
\newcommand{\ps}{\mathbf{p}} 

\newcommand{\z}{{z}}  

\newcommand{\cl}{{c}} 

\renewcommand{\c}{{ctx}} 
\renewcommand{\t}{ {} } 

\newcommand{\test}{{}^*} 

\newtheorem{theorem}{Theorem}

\def\our{ZEUS}

\title{\our{}: Zero-shot Embeddings\\ for Unsupervised Separation of Tabular Data}

\author{%
  Patryk Marsza{\l}ek$^{1,2}$ 
    \And
    Tomasz Ku\'smierczyk\thanks{Joint contribution to project conception, design, and research supervision.}$^{\:\  ,1}$
    \And 
    Witold Wydma\'nski$^{1,2}$
    \And
    Jacek Tabor$^{1}$
    \And 
    Marek \'Smieja$^{*,1}$\\
    \\
    \textsuperscript{1}Faculty of Mathematics and Computer Science, Jagiellonian University, Kraków, Poland \\
    \textsuperscript{2}Doctoral School of Exact and Natural Sciences, Jagiellonian University, Kraków, Poland \\ \\
\texttt{\{t.kusmierczyk,witold.wydmanski,jacek.tabor,marek.smieja\}@uj.edu.pl} \\
\texttt{patryk.marszalek@doctoral.uj.edu.pl}
}

\begin{document}

\maketitle

\begin{abstract}
Clustering tabular data remains a significant open challenge in data analysis and machine learning. Unlike for image data, similarity between tabular records often varies across datasets, making the definition of clusters highly dataset-dependent. Furthermore, the absence of supervised signals complicates hyperparameter tuning in deep learning clustering methods, frequently resulting in unstable performance. To address these issues and reduce the need for per-dataset tuning, we adopt an emerging approach in deep learning: zero-shot learning. We propose \our{}, a self-contained model capable of clustering new datasets without any additional training or fine-tuning. It operates by decomposing complex datasets into meaningful components that can then be clustered effectively. Thanks to pre-training on synthetic datasets generated from a latent-variable prior, it generalizes across various datasets without requiring user intervention. To the best of our knowledge, \our{} is the first zero-shot method capable of generating embeddings for tabular data in a fully unsupervised manner. 
Experimental results demonstrate that it performs on par with or better than traditional clustering algorithms and recent deep learning-based methods, while being significantly faster and more user-friendly.
\end{abstract}

\section{Introduction}

Clustering remains a fundamental yet challenging task in unsupervised learning. It is particularly hard for tabular data, which inherently lacks the structured spatial or semantic properties of images or texts. Unlike for image clustering, where intrinsic visual similarities can guide cluster formation, defining meaningful similarities in tabular data is highly dataset-specific, complicating the generalization of clustering methods across diverse applications. Recent developments leveraging deep learning have demonstrated promise in generating richer representations for clustering tasks. However, these methods frequently suffer from instability due to their sensitivity to hyperparameter selection, a challenge exacerbated by the absence of supervised signals to guide optimization. Consequently, practitioners working with tabular data often resort to simpler, classical algorithms like k-means~\cite{macqueen1967some}, despite their limited capacity for capturing complex underlying data structures, simply to avoid extensive manual tuning.

We address these challenges with \our{} -- a \textbf{z}ero-shot transformer-based model for \textbf{e}mbedding new tabular datasets in a form convenient for \textbf{u}nsupervised \textbf{s}eparation (=clustering), without the need for additional fine-tuning. Given a new dataset, \our{} returns its transformed representation, where clusters can be easily discovered using simple methods, like k-means. Since it works as a zero-shot learner, it significantly reduces hyperparameter tuning complexity and computation time, enabling effective clustering in seconds.
It is a plug-and-play solution that requires no fine-tuning and runs in a single forward pass for new datasets.

Inspiration for \our{} stems from the Prior-data Fitted Networks (PFNs)~\cite{tabpfn}, a recently introduced framework highlighting the potential of in-context learning for tabular data. PFNs operate by feeding a transformer~\cite{transformer} contexts representing complete tasks, i.e., training data along with query samples for which the model predicts labels. Despite their impressive performance, they are limited to supervised problems. \our{} extends this basic idea for unsupervised tasks by addressing the two key challenges of how to: (1) generate prior synthetic data with clear but non-trivial clustering structures for pre-training; and (2) encode prior clustering knowledge for new unlabeled datasets. 
Additionally, in contrast to TabPFN, \our{} is a zero-shot model and during inference does not rely on any context labels.

\begin{figure}[t]
    \centering
    \includegraphics[width=0.975\linewidth]{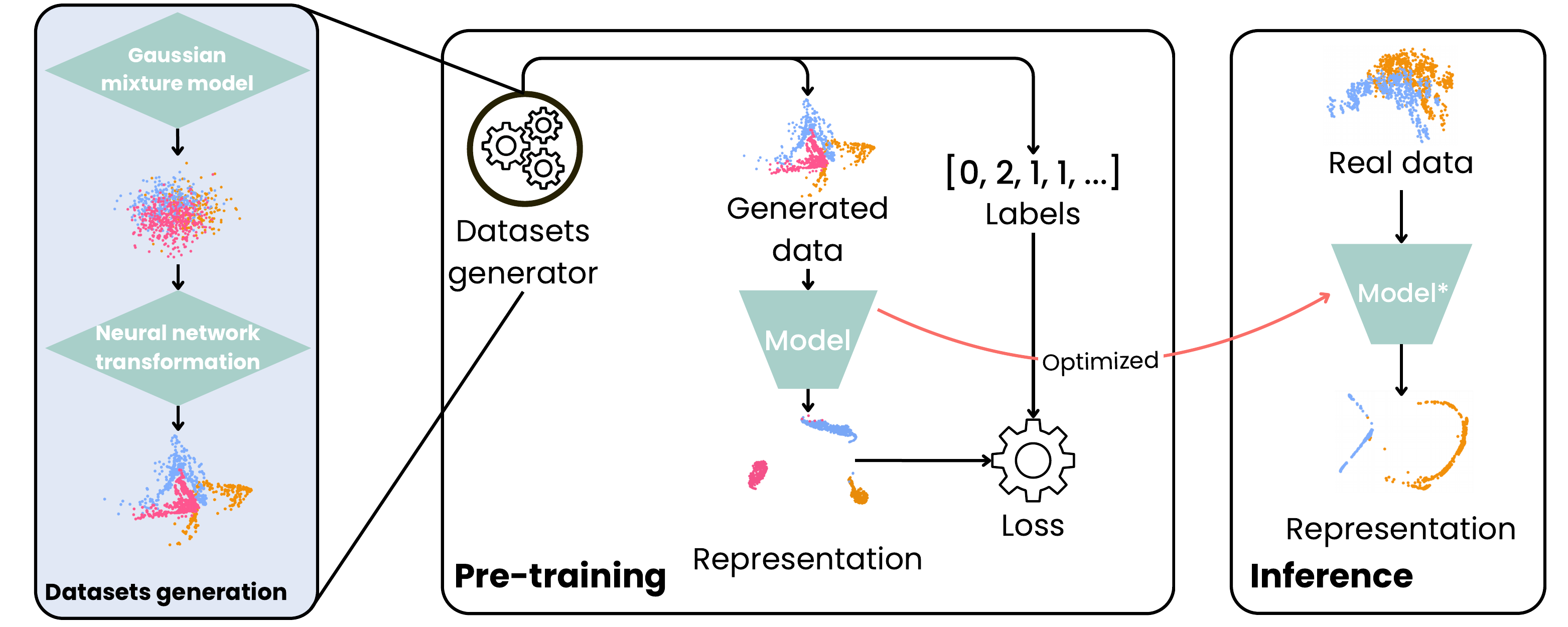}
    \caption{Schematic characterization of \our{}: (left) synthetic datasets generation; (middle) pre-training on datasets with known labels; (right) deployment of a frozen model for real-world tasks.
    }
    \label{fig:scheme}
\end{figure}

\begin{figure}[t]
    \centering
    \includegraphics[width=0.9\textwidth]{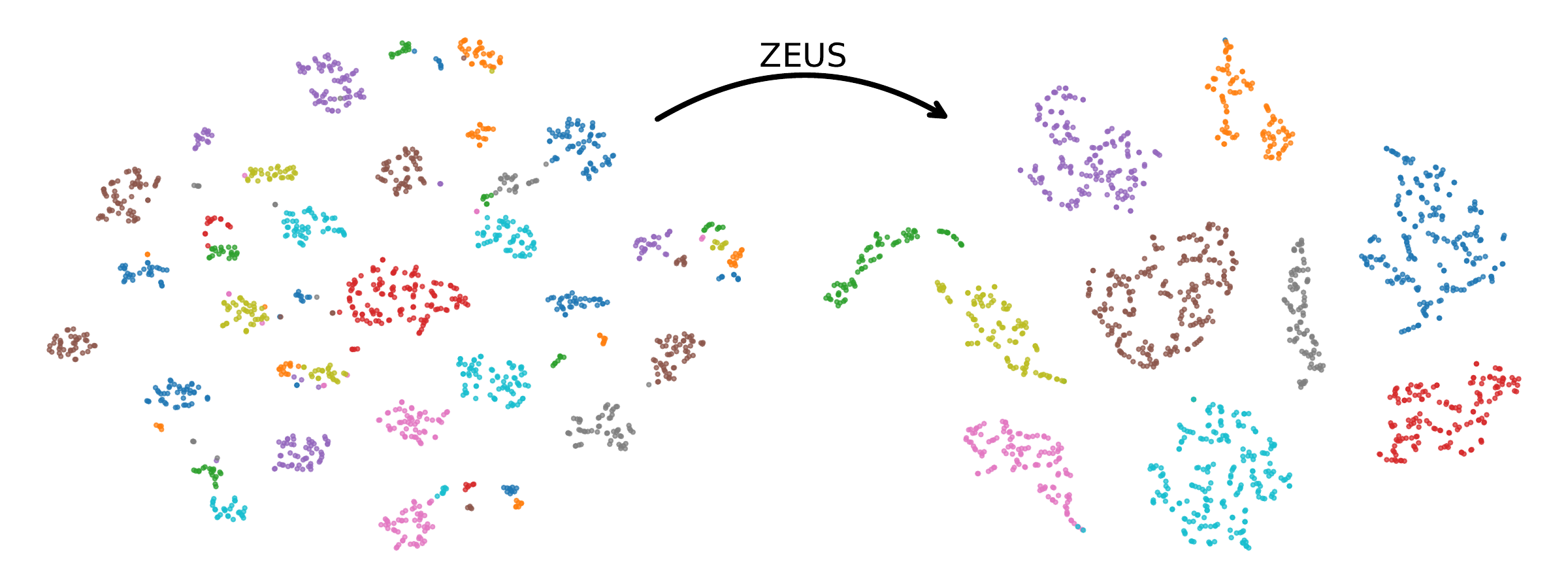}
    \caption{t-SNE visualization for a sample synthetic dataset. The representation from \our{} (right panel) significantly improves consistency with ground-truth classes and reveals a clearer data structure.}
    \label{fig:tsne_visualization_1}
\end{figure}

Unlike traditional methods that optimize arbitrary heuristics (e.g., DEC~\cite{xie2016unsuperviseddeepembeddingclustering}), \our{} approaches clustering by learning how to invert data generation processes. In particular, it is pre-trained on synthetic datasets to learn how to infer cluster assignments. The datasets have known latent structures, 
which enables supervised guidance for the training, and by generating diverse 
datasets 
we supply it with the prior knowledge about what can constitute a possible clustering structure, 
enabling effective generalization to new real-world datasets during inference.
Figure~\ref{fig:scheme} illustrates the key concepts of the method while \Cref{fig:tsne_visualization_1} presents representations generated by \our{}.

Experiments demonstrate that \our{} consistently matches or surpasses both classical clustering algorithms and recent deep learning-based approaches, establishing it as a powerful and practical tool for the unsupervised analysis of tabular data. Beyond its strong empirical performance, we also provide theoretical justification by showing that \our{} fits the framework of Prior-data Fitted Networks, thereby reinforcing its theoretical soundness (see \Cref{sec:pfn}).

We supplement the paper with an appendix that includes background on prior-data fitted networks, specifics of the synthetic data generation process, details of the experimental results, and additional experimental studies. The code used in this paper is available at \href{https://github.com/gmum/zeus}{https://github.com/gmum/zeus}.

\section{Pre-training representation for clustering}

In this section, we introduce our approach to zero-shot representation pre-training for clustering. First, we characterize the proposed probabilistic model and its training objective. Then, we explain the process of generating synthetic datasets for pre-training. Finally, we highlight the connection between our model, Bayesian learning and the framework for prior-data fitted networks.

\subsection{\our{} method}
\label{sec:method}

\paragraph{Zero-shot pre-training.}

Traditional machine learning models $\Model_\theta$ are trained by optimizing parameters $\theta$ to solve a specific \emph{fixed} task, such as classification or clustering, over a fixed dataset. 
On the other hand, \emph{in-context learning} (ICL) \cite{garg2023transformerslearnincontextcase, lee2023supervised, han-etal-2023-understanding} enables \emph{pre-trained models} $\Model_{\theta^\ast}$ to adapt to new tasks or datasets without ever updating the pre-trained parameters $\theta^*$.
Similarly, \emph{zero-shot learning} allows models to address problems they were never explicitly trained on, without requiring any examples of the target task. Unlike ICL, which adapts through demonstration examples, zero-shot approaches rely entirely on knowledge encoded during pre-training. The pre-training of zero-shot models typically involves massive and diverse datasets that encourage the model to learn rich, generalizable representations of underlying patterns and relationships.
This is usually done using 
\emph{transformers}, as they both can process set-valued inputs (e.g., accept a whole dataset at once) and model complex dependencies.

Our approach involves pre-training a transformer $\Model_{\theta}$ on a diverse collection of datasets $\D_\t$ such that $\D_\t := (\xs_\t, \ys_\t)$, where $\ys_\t=\{y_i\}$ denotes ground-truth labels (e.g., clusters) and $\xs_\t=\{x_i\}$ are feature vectors.
A loss $\Loss$ encourages $\Model_{\theta}$ to assign accurate labels $\ys_\t$ to all data points in the target datasets $\D_\t$:
\begin{equation}
    \label{eq:argmin}
    \theta^{\ast}=\arg\min_{\theta}\;
    \mathbb{E}_{p(\D)}\bigl[\Loss\bigl(\ys,\,
    \Model_{\theta}(\xs_\t)\bigr)\bigr] 
    \approx \arg\min_{\theta}\;
    \frac{1}{N_\D}  \sum_{\D_\t \sim p(\D)} \Loss\bigl(\ys_\t,\,
    \Model_{\theta}(\xs_\t)\bigr),
\end{equation}
where $p(\D)$ is a process generating \emph{prior datasets} which \emph{implicitly} characterize a space of possible target tasks (here approximated with $N_\D$ Monte-Carlo samples).
{Note, the loss $\Loss$ \emph{may factorize} over individual points as $\Loss(\ys_\t, \hat \ps_\t) = \sum_i \ell(y_i, \hat p_i)$, where $\hat p_i$ denotes a probabilistic prediction for an individual input $x_i$, but unlike for classification, for tasks such as clustering or anomaly detection, the predictions $\hat p$ must be made \emph{jointly} for all inputs. Hence, $\Model_\theta(\xs_\t)$ can not be decomposed into predictions for individual inputs $x_i$.} 
Furthermore, in contrast to typical supervised ICL, training labels are not included in the context, as they are unavailable during inference. This makes the unsupervised learning problem significantly more challenging, since we cannot explicitly guide the model toward the desired structure in a given dataset.


\paragraph{Representation learning for probabilistic clustering.}
\label{sec:loss}
%
We pre-train an encoder $ \Model_\theta $ to map inputs $\xs_\t$ to representation vectors 
$
\z(x_i) = f_\theta(\xs_\t)_i \in \mathbb{R}^D =: S. 
$
We aim at positioning data points in $S$, so that they naturally form probabilistic clusters, and
our goal is to make the encoder \emph{find better representations for target tasks}. 
%
%
This objective
we frame as maximizing the log-likelihood of correct cluster assignments. 
Since the ground-truth cluster assignments $y_{i}$ are available during pre-training, this is expressed as:
\begin{equation}
    \label{eq:loss_ce}
    \Loss_{prob}  =  - \sum_{i} \log p_{y_{i}}(x_{i}).
\end{equation}
For specifying $p_{y}(x)$, we draw inspiration from Gaussian Mixture Models (GMMs)~\cite{dempster1977maximum,hansen1982large}, where
each cluster $k$ is represented by a Gaussian distribution centered at $\cl_{k}$, and data points are probabilistically assigned to clusters through probabilities $p_{k}(x_{i})$, based on their proximity to these centroids.
Consequently,
alongside the encoded representations, we need to define a set of $ K $ cluster centroids (e.g., prototypical representations) $ \{\cl_k\}_{k=1}^K $, which reside in the same representation space $ S $.  
We then \emph{jointly} optimize $\Loss$ w.r.t $ \theta $ and $ \{\cl_k\} $ such that the representations $ \z(x_i) $ become structured in a way undisclosing \emph{underlying probabilistic structures} in datasets.

We aim for higher probabilities $p$ for points closer to these centroids. To achieve this, we first define scores 
$
\alpha_{k}(x_{i})  =  - \| \z(x_{i}) - \cl_{k} \|^2
$
to quantify the compatibility between representations and centroids $\cl_k$, and then transform them into cluster membership probabilities using softmax, yielding \emph{soft} cluster assignments
$
p_{k}(x_{i})  =  \frac{\exp(\alpha_{k}(x_{i}))}{\sum_{j  =  1}^K \exp(\alpha_{j}(x_{i}))}.
$
Note that the ground-truth cluster assignments $y_{i}$ are known during pre-training, and hence, the maximum likelihood centroids can be straightforwardly estimated as
$
\hat \cl_{k}  =  \frac{1}{N_{k}}\sum_{\{i : y_{i}  =  k\}} \z(x_{i}),
$
where $N_{k}$ denotes the number of data points assigned to the $k$-th cluster. 
Following this formulation,  the optimization needs to be performed only w.r.t the parameters $\theta$ of the neural network $\Model_\theta$, as the centroids are already specified by them.

Besides the centroids, we additionally consider another concept from GMMs, namely cluster priors~${\pi}_{k}$.
Similar to centroids, the prior cluster frequencies $\hat{\pi}_{k}$ can be estimated from the training labels as the proportions of examples in the $k$-th cluster, and then incorporated directly into the probability calculation, resulting in the final form of our probabilistic scoring:
\begin{equation}
\label{eq:prob_with_priors}
    p_{k}(x_{i})  =  \frac{\hat{\pi}_{k} \exp(\alpha_{k}(x_{i}))}{\sum_{j  =  1}^K \hat{\pi}_{j} \exp(\alpha_{j}(x_{i}))}, \quad \textit{where} \quad \alpha_{k}(x_{i})  =  - \| \z(x_{i}) - \hat \cl_{k} \|^2.
\end{equation}

\begin{remark}
\our{} cluster assignments correspond to the membership probabilities in GMMs
defined as:
$
r_{ik}  =  \frac{\pi_{k} \mathcal{N}(\z(x_{i})|\cl_{k}, \Sigma_{k})}{\sum_{j  =  1}^K \pi_{j} \mathcal{N}(\z(x_{i})|\cl_{j}, \Sigma_{j})},
$
but with fixed diagonal covariance matrices $\Sigma_{k}  :=  I$, effectively forcing the encoder to learn circular clusters rather than elliptical ones. 
\end{remark}

\paragraph{Inference.} 
The structural relations learned during pre-training are leveraged during inference when the representations $\z$ produced by the model $\Model_{\theta^{\ast}}$ are used to create a predictive distribution for new, previously unseen inputs $\xs_\test$ as
$\hat \ps_\test := p\bigl(\ys_\test \mid \Model_{\theta^{\ast}}(\xs_\test)\bigr).$
At this point, the labels $\ys_\test$ are unknown and therefore cannot be used to estimate the centroids nor the priors. Hence, to structure the obtained representation into clusters, we use a traditional learning algorithm, for example, similar to GMMs relying on Expectation-Maximization or simply k-means.

\paragraph{Regularization.}
\label{sec:regilarization}
Although theoretically $\Loss_{prob}$ is sufficient to build a clustering representation, we experimentally verified that the introduction of additional regularizers further improves the predictions in the inference (see \Cref{sec:regularization}).

First, we aim to ensure that the representations $ \z(x_i) $ associated with a particular cluster are compactly distributed around their corresponding centroid to enhance intra-cluster cohesion. We define a point concentration regularizer that explicitly minimizes the distance between representations and the centroids, analogous to the k-means objective of minimizing within-cluster sum of squares:
\begin{equation}
    \Loss_{cp} = \sum_k \sum_{i: y_i=k} \alpha_k(x_i).
\end{equation}

Second, to prevent the cluster centroids $ \cl_k $ from collapsing towards similar points in the space, we add a centroid separation regularizer.
We achieve this by maximizing the sum of squared distances between all distinct pairs of centroids. However, to avoid this term dominating the loss if centroids were pushed infinitely far apart, we cap the contribution of each pair's squared distance at a predefined threshold $ T $. The term to be minimized is thus:
\begin{equation}
    \Loss_\textit{sep} = -\sum_{k=1}^K \sum_{j=k+1}^K \min (\|\hat \cl_k - \hat \cl_j\|^2, T).
\end{equation}

The \emph{final loss} combines the main clustering objective with two regularization terms as
\begin{equation}
\Loss = \Loss_{prob} + \lambda_{cp} \Loss_{cp} + \lambda_{sep} \Loss_{sep}
\end{equation}
where the hyperparameters $ \lambda_{cp} \ge 0 $ and $ \lambda_{sep} \ge 0 $ promote point concentration and control the relative importance of enforcing centroid separation, respectively.
We used simply $ \lambda_{cp} = \lambda_{sep} = 1 $.

\subsection{Prior data for pre-training \our{}}
\label{sec:prior}
The key component of \our{} is the data-generating prior $p(\D)$. As it primarily affects the generalization ability of the pre-trained model $\Model_{\theta^*}$, it must be designed to cover diverse data distributions and various cluster configurations. 
In particular, we construct it as a latent variable model (LVM), assuming that each dataset $\D$ is sampled from a $K$-component mixture of distributions. Formally, the probability of a data point $x \in \D$ under this model is
$$
p(x) = \sum_{k=1}^K p(y = k) \, p(x \mid y = k),
$$
where $p(y)$ is a categorical distribution, and $p(x \mid y = k)$ represents the $k$-th (continuous) component. Although the categories $y$ remain latent for real datasets, they are known during synthetic data generation and can serve as ground-truth labels -- a property integral to pre-training \our{}. 
This procedure, coupled with Theorem~\ref{theorem} (in Appendix) ensures that a sufficiently expressive model trained on synthetic datasets can handle  real datasets drawn from arbitrary mixtures of distributions.

The number of categories $K$ we sample uniformly between 2 and  10, and the observations $x$ are sampled from multivariate Gaussian distributions. To control the complexity of datasets, we introduce a constraint (Eq.~\ref{eq:gaussian_constraint} in Appendix) that ensures a sufficient separation between each pair of components. For each component, we generate between $50$ and $800$ samples.
Since real data rarely consist of Gaussian clusters, we additionally transform the data points from each category using randomized ResNet-like neural networks to produce more realistic cluster shapes. 
We selected ResNets due to their properties, as they can define invertible transformations ~\citep{behrmann2019invertibleresidualnetworks}, and this ensures that clustering structures will be preserved in outputs. Finally, we append a certain fraction of categorical features in one-hot encoding scheme to selected datasets, which is followed by an optional PCA reduction to keep the data dimension at the requested maximum level. 
Complete details of the prior-data generating process can be found in Appendix \ref{sec:gen}. Theoretical justification of this procedure is presented in Appendix~\ref{app:theory}.

The above strategy for generating clustering datasets illustrates the key idea behind \our{}. Instead of clustering each dataset individually using an arbitrary loss criterion (as in $k$-means or DEC), \our{} learns to perform clustering by inverting the data generation process.  
Although our synthetic datasets are limited by the selected family of distributions specified above, the proposed paradigm for zero-shot unsupervised learning for clustering is general.

\subsection{Relation to Bayesian learning and Prior-Data Fitted Networks}
\label{sec:pfn}

The recently introduced framework of \emph{Bayesian inference through transformers} demonstrates that neural networks pre-trained on synthetic datasets implicitly approximate Bayesian inference without explicitly computing posterior distributions \cite{bayes_in_context, tabpfn}. 
By framing the approach described in Section~\ref{sec:method}  as a Prior-Data Fitted Network (PFN), we show \our{} implicitly performs approximate Bayesian averaging. 

Given a prior distribution $p(\D)$ over (synthetic) datasets $\D$, a PFN parameterized by $\theta$ is trained to minimize the negative log-likelihood of predicting held-out labels within datasets sampled from this prior. The associated loss function is defined as:
$
\Loss_{\textit{PFN}}(\theta)  =  \mathbb{E}_{\D_\c\cup \{(x,y)\}\sim p(\D)}[-\log q_{\theta}(y|x,\D_\c)]. 
$
Minimizing the Prior-Data Negative Log-Likelihood is then equivalent to minimizing the expected Kullback–Leibler divergence between the network's predictive distribution and the true Posterior Predictive Distribution (PPD) $
p(y|x,\D_\c)  =  \int_{\Phi} p(y|x,\phi)p(\D_\c|\phi)p(\phi)d\phi
$ (see Corollary 1.1 in \cite{bayes_in_context}). 
A pre-trained  PFN approximates this integral implicitly, yielding a distribution $q_{\theta}(y|x,\D_\c)$ directly from forward propagation of the network.
In \Cref{appendix:pfn}, a more detailed explanation of PFNs was provided.

\our{} instead of directly outputting $q_\theta(y|x,\D_\c)$,  maps inputs $x$ to latent representation vectors $\z(x)$.
The probabilistic assignments $p_y(x)$ are then constructed from these vectors according to Eq.~\ref{eq:prob_with_priors}. This equation specifies a PPD for inference, but also defines a probability mass function for training:
\begin{remark}
Eqs.~\ref{eq:argmin}, \ref{eq:loss_ce}, and \ref{eq:prob_with_priors} constitute a valid Prior-Data Negative Log-Likelihood, equivalent to Eq.~(2) from \cite{bayes_in_context} with $\D_\c=\emptyset$.
\end{remark}
This follows by mapping $p_y(x) := q_\theta(y|x,\emptyset)$ and noting that Eq.~\ref{eq:loss_ce} corresponds to the cross-entropy between the true labels $\{y\}$ and the probabilistic assignments $p_y(x)$.
Intuitively, this formulation reinterprets probabilistic clustering with known labels as a classification task.
Then, pre-training the transformer by minimizing the cross-entropy loss over prior-generated datasets remains identical to PFNs, and thus, we can conclude that \our{} \emph{implicitly learns a Bayesian approximation through prior fitting}.

Our method, however, deviates from traditional PFNs by imposing an explicit  mixture-like structure on latent representations (Eq.~\ref{eq:prob_with_priors}), unlike more general PFNs:
\begin{remark}
By enforcing the clustering structure, \our{} may impose stronger assumptions on the PPD compared to vanilla PFNs. This structure might be suboptimal for classification tasks and the true PPD may not belong to the family of attainable solutions.
\end{remark}
As explained above, our clustering-based extension is theoretically sound, nonetheless, the assumption could limit representational expressivity compared to the more flexible transformers employed for the original PFNs~\cite{tabpfn}. On the other hand, the enforced structures shall be more appropriate for unsupervised tasks.


\section{Experiments}
\label{sec:experiments}
This section presents an experimental evaluation of the clustering performance of our method. Due to space constraints, further results are provided in Appendix~\ref{aappendix:additional_exp}.

\subsection{Experimental setup}
\label{sec:setup}

\textbf{Model architecture:} 
\our{} relies on a transformer architecture similar to TabPFN~\citep{tabpfn}. It consists of 12 attention blocks, each with 6 heads and a token dimension of 512, with GeLU activation employed. Following the TabPFN design, each data point is first embedded using a linear transformation and then passed as a token to the transformer. For the unsupervised setting, no label embeddings are ever created or presented to the model. Similarly, for the zero-shot case, no query set is used either, and consequently, attention is computed solely over the support set. Finally, unlike TabPFN, we omit any additional MLP decoder after the transformer.

\textbf{Pre-training:}
In the pre-training phase, we sample datasets from the mixture of Gaussians and transformed mixtures in equal proportions (1:1), generating 1000 unique dataset batch samples for each epoch.  For training, we employ the Adam optimizer along with a cosine learning rate scheduler with warm-up, using a learning rate of 2e-5. 
The plot illustrating model improvements during the pre-training process is available in \Cref{appendix:pre-training graphs}.

\textbf{Inference:}
During inference, preprocessing of each dataset involves standardizing numerical features, followed by scaling them to the range $[-1, 1]$, whereas categorical features are transformed using one-hot encoding. The input size of our model is fixed to 30. For datasets with lower dimensionality, we pad the missing positions with zeros, while for higher-dimensional datasets, we reduce the number of input features via Principal Component Analysis (PCA).
Unless stated otherwise, at inference we use k-means applied to the normalized (=scaled to $[-1,1]$) transformer output in order to obtain clusters from our learned representation.



\textbf{Datasets:} For evaluation, we consider three groups of datasets: real datasets from OpenML~\cite{OpenML2025} (\emph{Real}), synthetic mixtures of Gaussians (\emph{Syn. Gauss.}), and synthetic mixtures of Gaussians transformed by ResNet-like neural networks (\emph{Syn. Transf.}). Both types of synthetic datasets are augmented with categorical variables. The process of generating synthetic datasets is described in \Cref{sec:prior}.

Each dataset contains at most 2000 samples as per model design and due to memory limitations. The study covers 34 real datasets, selected based on their clustering feasibility, defined as ARI $\geq 0.4$ achieved by at least one of the methods. Additionally, 20 synthetic datasets of each type were generated from the same prior as in the pre-training phase, but with a different random seed to ensure a fair comparison. Detailed statistics of the datasets, such as the number of numerical and categorical features, are provided in Appendix~\ref{appendix:statistics}.

\textbf{Baselines:}
\label{sec:baselines}
We compare \our{} against a wide spectrum of state-of-the-art clustering methods used for tabular data. It includes $k$-means (KM), Gaussian Mixture Model (GMM), and deep-learning methods based on autoencoder (AE) architectures, including DEC~\cite{xie2016unsuperviseddeepembeddingclustering}, IDEC~\cite{guo_improved_2017}, IDC~\cite{idc}, and G-CEALS~\cite{rabbani2025deep}. We additionally consider $k$-means and GMM performed in the autoencoder latent space (respectively referred to as AE-KM and AE-GMM), along with $k$-means applied to representations obtained from TabPFN-Unsupervised~\cite{hollmann2025accurate} (TabPFN) and SCARF~\cite{scarf}, to assess the clustering quality of different feature representations.


For AE-based baselines, we employ the standard configuration used in the prior literature, i.e., an architecture comprising of hidden layers with sizes [500, 500, 2000]. We  used the latent dimension of 20, which we verified experimentally  as the best value. Further details on the hyperparameters and code repositories for the baselines are provided in Appendix \ref{appendix:baselines}.

\textbf{Reporting:}
In the main text, we report only the results aggregated for each group of datasets (\emph{Real}, \emph{Syn. Gauss}, \emph{Syn. Transf.}), while detailed results for individual datasets can be found in \Cref{sec:exp_app}. For the reader's convenience, we \textbf{bold} the best results and \underline{underline} the second-best ones.

\subsection{How effective is \our{} for clustering?}
\label{sec:quantitative_results}

To evaluate the performance of the clustering methods, we employ a standard evaluation procedure, where clusters identified by models are expected to correspond to undisclosed ground-truth classes. 
We use the Adjusted Rand Index (ARI) for quantitative evaluation, and 
to improve readability, we scale the ARI values by a factor of 100, where 100 indicates perfect clustering and values near 0 represent random grouping. Table~\ref{tab:comp_ari} presents a summary of ARI scores averaged over 5 random seeds and all datasets within each dataset group. Additionally, Table~\ref{tab:comp_rank} displays the average rankings for each of the methods.



\begin{table}[t]
    \caption{Clustering quality (ARI) of \our{} versus competing methods (higher is better).}
    \centering
    \scriptsize
    \begin{tabular}{lccccccccccc}
        \toprule
        & KM & GMM & AE-KM & AE-GMM & DEC & IDEC & IDC & G-CEALS & TabPFN & SCARF & \our \\
        \midrule
        Real & 55.54 & 48.49 & 51.43 & 53.56 & \underline{55.93} & 54.57 & 52.28 & 40.37 & 31.32 & 26.95 & \textbf{57.43} \\
        Syn. Gauss. & \textbf{89.90} & 76.93 & 81.26 & 81.40 & \underline{89.35} & 82.57 & 66.43 & 62.84 & 55.97 & 8.32 & 89.03 \\
        Syn. Transf. & 75.04 & 75.88 & 60.45 & 71.29 & \underline{79.94} & 61.26 & 66.78 & 49.17 & 15.66 & 2.48 & \textbf{86.33} \\
        \bottomrule
    \end{tabular}
    \label{tab:comp_ari}
\end{table}

\begin{table}[t]
    \caption{Average rank of the methods used in the benchmark (lower is better).}
    \centering
    \scriptsize
    \begin{tabular}{lccccccccccc}
        \toprule
        & KM & GMM & AE-KM & AE-GMM & DEC & IDEC & IDC & G-CEALS & TabPFN & SCARF & \our \\
        \midrule
        Real & \underline{4.69} & 5.65 & 5.72 & 5.24 & \underline{4.69} & 5.01 & 5.62 & 8.18 & 8.22 & 8.85 & \textbf{4.13} \\
        Syn. Gauss. & \textbf{2.65} & 3.65 & 5.65 & 5.60 & 3.23 & 5.70 & 7.95 & 8.90 & 8.75 & 11.00 & \underline{2.92} \\
        Syn. Transf. & 4.80 & 3.50 & 6.85 & 4.50 & \underline{3.20} & 6.35 & 6.05 & 7.80 & 9.75 & 11.00 & \textbf{2.20} \\
        \bottomrule
    \end{tabular}
    \label{tab:comp_rank}
\end{table}

\begin{wrapfigure}{r}{0.525\textwidth}
  \begin{center}
\includegraphics[width=0.525\textwidth]{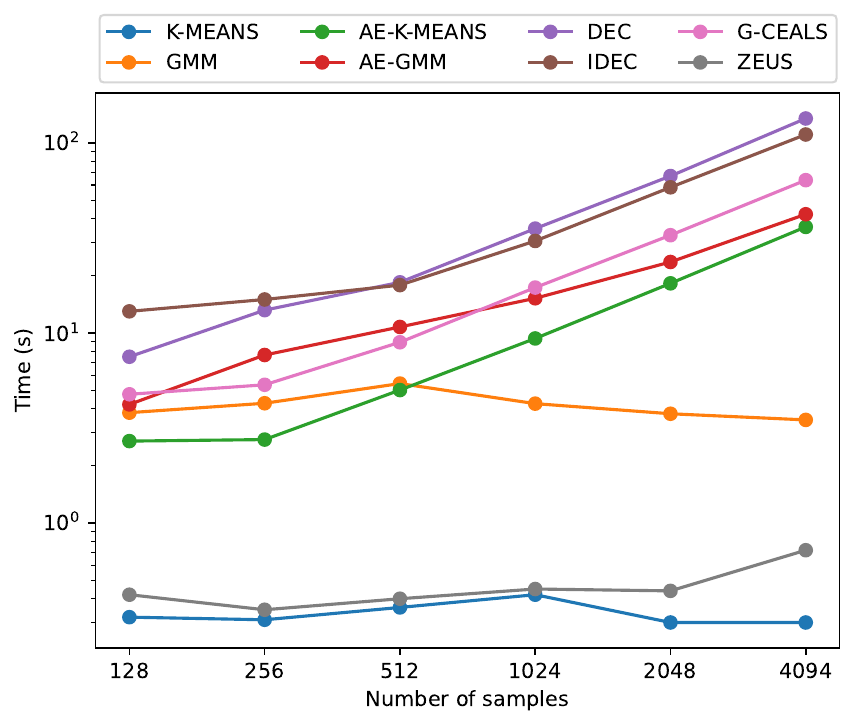}
  \end{center}
  \caption{Clustering time~vs.~input size.}\label{fig:time}
\end{wrapfigure}

We observe that \our{} achieves competitive performance for all three groups of datasets. In particular, it achieves the best average ARI and the best rank for both the OpenML and Synthetic Transformed datasets, e.g., for the most challenging clustering scenarios. Notably, \our{} outperforms the second-best method, DEC, by more than 6 percentage points, and the classical baselines by over 10 percentage points on the Synthetic Transformed datasets. The results on the Real datasets further demonstrate that \our{} effectively generalizes the knowledge acquired during the pre-training stage on synthetic data to general data distributions. For the simplest Gaussian datasets, \our{} ends up in top-3, with results only slightly below those of k-means and DEC. In terms of average rank it gets the second place.

Among the AE-based methods, DEC performs the best, while G-CEALS consistently demonstrates the weakest performance across all datasets. K-means and GMM, as representatives of the classical methods, perform reasonably well. In particular, k-means consistently lands among the best-performing methods, which justifies its broad adoption among practitioners, despite being one of the most basic approaches. 
On the other hand, approaches such as SCARF or TabPFN 
are well-suited for tasks like classification,
thanks to their strong performance in feature extraction. 
However, without additional regularization during training or specialized post-hoc fine-tuning (such as with DEC), effective separation of their representations is difficult.
The detailed results, including scores for individual datasets corresponding to the averages in Tables~\ref{tab:comp_ari} and~\ref{tab:comp_rank}, can be found in Appendix~\ref{appendix:extended_results_ari}

Finally, \Cref{fig:time} illustrates how the examined clustering methods scale with the increasing number of input data points. \our{} maintains almost constant time while being only slightly slower than the basic k-means. 
It shows that the overhead from creating the representations by \our{} is minimal.
On the other hand, the remaining deep clustering algorithms require significantly more time and scale poorly with the increasing input size.

\subsection{Are \our{}'s predictions calibrated?}
\label{sec:brier_exp}

For applications where uncertainty quantification is as crucial as the clustering decisions themselves, assessing calibration is particularly relevant. Having demonstrated the strong performance of \our{} for hard clustering, we now examine whether its probabilistic foundations yield well-calibrated soft assignments. We compare it against competing approaches using the Brier score, which
measures the accuracy of predicted probabilities. Unlike the previously used ARI, it penalizes both incorrect cluster assignments and \emph{poorly calibrated confidence scores}, thereby providing a more comprehensive evaluation. 

The Brier score is a supervised metric, meaning its direct computation for unsupervised clustering tasks is not straightforward. However, when ground-truth classes are available and the number of clusters matches the number of classes, a one-to-one mapping between clusters and classes can be established 
using the Hungarian algorithm \citep{kuhn1955hungarian,kuhn1956variants}, which aims to maximize the total agreement between cluster-class pairs. The cost matrix $A$ for this assignment problem is $A_{jc} = \sum_{i=1}^{N} p_{ij} \cdot Y_{ic}$, where $Y_{ic} = 1$ if data point $i$ belongs to class $c$, and $Y_{ic} = 0$ otherwise.

Table~\ref{tab:comp_brier} reports the average Brier score computed over 5 seeds  for all datasets in respective groups.
For the analysis, we used the vanilla variant of \our{}, which is paired with the GMM clustering since k-means does not provide soft assignments. The covariances were constrained to be identities as implied by Eq.~\ref{eq:prob_with_priors}. 
All competing baselines, except for k-means, provide probabilistic cluster assignments, making the Brier score calculation straightforward for them. For k-means, we used one-hot encoding to represent its assignments as probabilities.

\begin{table}[t]
    \footnotesize
    \caption{
Soft clustering quality of \our{} vs. baselines, measured by Brier score (lower is better). 
    }
    \centering
    \begin{tabular}{lcccccccc}
        \toprule
        & KM & GMM & AE-KM & AE-GMM & DEC & IDEC & G-CEALS & \our \\
        \midrule
        Real & 0.4366 & 0.4799 & 0.4643 & 0.4679 & 0.3941 & \textbf{0.3671} & 0.4722 & \underline{0.3817} \\
        Syn. Gauss. & \textbf{0.0970} & 0.3110 & 0.2073 & 0.2484 & 0.3943 & 0.2308 & 0.3946 & \underline{0.1269} \\
        Syn. Transf. & 0.2803 & \underline{0.2566} & 0.4535 & 0.3140 & 0.4638 & 0.3892 & 0.4951 & \textbf{0.1796}  \\
        \bottomrule
    \end{tabular}
    \label{tab:comp_brier}
\end{table}

\our{} demonstrates outstanding performance, 
achieving the best results for the Synthetic Transformed datasets while
ranking second on both the OpenML and Synthetic Gaussian collections. 
The benefits of using \our{} representations are especially visible in comparison with the vanilla GMM. Although the GMM achieves the second-best score on transformed data, its performance across the remaining datasets is merely modest. Among the remaining baselines, IDEC exhibits an interesting pattern: despite relatively weak clustering performance in Table~\ref{tab:comp_ari}, it achieves the top position  on real datasets according to the  Brier score and shows marked improvement on synthetic data. DEC displays the opposite tendency, with calibration results significantly inferior to its strong ARI performance. While G-CEALS remains generally weaker than other methods, its calibration performance consistently exceeds its clustering results presented in Table~\ref{tab:comp_ari}. The K-means algorithm, both in its standard implementation and when applied to autoencoder embeddings, yields impressive scores on Gaussian-categorical mixture datasets -- likely due to the inherent separability of these data structures, which allows even simple one-hot probability estimates to produce well-calibrated predictions.

\subsection{How helpful is regularization for \our{}?}
\label{sec:regularization}

Representations obtained by minimizing the basic loss $\Loss_{prob}$ can be further improved by regularizing the optimization process to structure the latent representations. In Table~\ref{tab:comp_ablations}, we examine how different combinations of the $\Loss_{prob}$ loss with the regularizers $\Loss_{cp}$ and $\Loss_{sep}$ affect the final performance across various datasets. To ensure a fair comparison, all models were evaluated using a data generator with fixed (identical for all) settings.

\begin{table}[ht!]
    \footnotesize
    \centering
    \caption{Regularisation impact, measured by ARI (higher is better).}
    \label{tab:comp_ablations}
    \begin{tabular}{lcccc}
        \toprule
         & $\Loss_{prob}$ & $\Loss_{prob} + \Loss_{sep}$ & $\Loss_{prob} + \Loss_{cp}$ & $\Loss_{prob} + \Loss_{sep} + \Loss_{cp}$ \\
        \midrule
        Real & 44.80 & \underline{51.60} & 48.65 & \textbf{57.43} \\
        Syn. Gauss. & 83.37 & 81.88 & \textbf{90.59} & \underline{89.03} \\
        Syn. Transf. & 79.85 & 79.29 & \textbf{88.58} & \underline{86.33} \\
        \bottomrule
    \end{tabular}
\end{table}

Our main model, which during pre-training incorporates both regularizers, exhibits a clear performance advantage on real datasets and consistently holds second place for the synthetic ones.
Table~\ref{tab:comp_ablations} shows that a model pre-trained using only 
$\Loss_{prob}$ and its variant with $\Loss_{sep}$ included are insufficient for properly separating the transformer's representations for the clustering task, as they both struggle with clustering the synthetic datasets encountered during pre-training.
On the other hand, the results for $\Loss_{prob} + \Loss_{cp}$ imply a positive impact by the compact loss component, $\Loss_{cp}$, especially for synthetic data, which, however, does not translate to strong performance on the OpenML benchmark.
We conclude that all three loss components are necessary for good performance, with $\Loss_{cp}$ being crucial for pre-training and $\Loss_{sep}$ important for real data.

This mismatch between performance on real and synthetic datasets suggests a potential prior misspecification, which is then mitigated by the regularizers. In particular, among the real datasets, there may be some that are not well explained by the data generation process used during pre-training. Hence, for future work, one may want to explore alternative data-generating priors that may be more appropriate for these outliers.

\subsection{What data-generating prior is optimal?}

As mentioned in Section~\ref{sec:prior}, constructing an appropriate data-generating prior $p(D)$ is crucial for maximizing performance and generalizability of \our{}. To validate this claim, we conduct an ablation study evaluating four prior designs: \emph{Gaussian + Categorical}, \emph{NN-transformed + Categorical}, \emph{Gaussian + NN-transformed}, and \emph{Gaussian + NN-transformed + Categorical} (the standard \our{} model). Each configuration name directly reflects the combination of priors used for pre-training (see Section~\ref{sec:gen} in the Appendix for further details). For a fair comparison, we keep the default hyperparameter setup unchanged, except for the prior itself. 

\begin{table}[!ht]
    \caption{ARI scores (averages; in rows) for \our{} variants pre-trained on different priors (columns).}
    \centering
    \footnotesize
    \begin{tabular}{lcccc}
        \toprule
        & Gauss. + Cat. & NN-transf. + Cat. & Gauss. + NN-transf. & Gauss. + NN-transf. + Cat.\\
        \midrule
         Real           & 40.59 & 50.90 & \underline{52.00} & \textbf{57.43} \\
        Syn. Gauss.    & \textbf{92.61} & \underline{89.90} & 75.25 & 89.03 \\
        Syn. Transf.   & 73.34 & \textbf{87.04} & 71.29 & \underline{86.33} \\
        \bottomrule
    \end{tabular}
    \label{tab:prior_ablation}
\end{table}

The results of the study are summarized in Table~\ref{tab:prior_ablation}.
As expected, the \emph{Gaussian + Categorical} model achieves top performance on the Synthetic Gaussian datasets, while the \emph{NN-transformed + Categorical} model excels for the Synthetic Transformed data. Notably, the latter also performs well on the Gaussian sets, as its prior is built upon transformed Gaussian mixtures. In contrast, the \emph{Gaussian + NN-transformed} model yields relatively poor average ARI scores across both synthetic benchmarks. This drop is primarily caused by a significant decline in ARI performance on the categorical datasets. Nonetheless, its average rank remains consistently competitive. Finally, for the OpenML datasets, the best performance is achieved when pre-training includes all three types of priors, underscoring the complementary importance of each prior in the training process.


\section{Related work}

\paragraph{Tabular data clustering.} Traditional clustering algorithms, like k-means \cite{macqueen1967some}, GMM \cite{dempster1977maximum}, or hierarchical methods \cite{murtagh2012algorithms}, have widespread applications across data mining, bioinformatics, customer segmentation, and anomaly detection. However, these methods often rely on predefined distance metrics and fail to capture complex, non-linear relationships, making them suboptimal for high-dimensional and heterogeneous tabular datasets. 

A pioneering Deep Embedded Clustering (DEC)~\cite{xie2016unsuperviseddeepembeddingclustering} improves the target data representation by training the autoencoder and computing soft assignments in its latent space via Student's $t$-distribution.
In a concurrent work \cite{yang2017towards}, the authors perform joint dimensionality reduction using AE and k-means clustering in the latent space. Then, improved DEC (IDEC)~\cite{guo_improved_2017} extends DEC by jointly optimizing reconstruction and clustering objectives, while spectral variants replace k-means steps with graph-based updates~\cite{8970928}. G-CEALS  \cite{rabbani2025deep} replaces the $t$–distribution assumption with multivariate Gaussian clusters.  DEPICT~\cite{dizaji2017deepclusteringjointconvolutional} attaches a softmax layer on top of an embedding network and trains with cross-entropy loss
to eliminate the assumption of an explicit distribution prior. Finally, IDC \cite{idc} 
predicts interpretable cluster assignments at the instance and cluster levels.

Most of these deep learning models require careful hyperparameter tuning and early stopping which is unrealistic in the fully unsupervised setting due to the lack of labels~\cite{9506051,sadeghi2024deepclusteringselfsupervisionusing}. Moreover, the optimization process has to be performed for each dataset, which is often time-consuming. Although multiple deep clustering approaches are currently in use, most of them are designed for texts or images \cite{znalezniak2023contrastive, cai2023semantic, li2024image} and cannot be directly adapted for tabular data due to the lack of a dominant neural architecture for heterogeneous tabular inputs \cite{grinsztajn2022tree, mcelfresh2023neural}.

\paragraph{Representation learning for tabular data.}
Self-supervised learning (SSL) has been transformative in domains like vision and language \cite{chen2020simple, grill2020bootstrap} but has struggled to show similar success in tabular data \cite{yoon2020vime, zhu2023xtab}. A large diversity of data and a lack of pre-defined correlation between features hinder the design of universal pretext tasks or augmentations, as well as the transfer learning between domains \cite{zhao2024comparison}. 
On the other hand, in-context and zero-shot learning~\cite{brown2020languagemodelsfewshotlearners} enable the use of a single pre-trained model for multiple tasks out of the box, without any additional tuning. In particular, 
TabPFN treats small tabular datasets as contexts consisting of features along with labels, and achieves state-of-the-art classification in a single forward pass~\cite{tabpfn}. Although theoretical analyses reveal that 
transformers can implicitly implement algorithms such as gradient descent \emph{in context}, their use is currently restricted to supervised problems~\cite{garg2023transformerslearnincontextcase}.


\section{Conclusion}

In this paper, we presented \our{}, a zero-shot transformer-based model that enables effective and efficient clustering of tabular data without the need for fine-tuning or extensive hyperparameter search. By pre-training on synthetic datasets with known latent structures, \our{} learns generalizable representations that help simple clustering algorithms to uncover meaningful structures. Our experiments show that \our{} consistently matches or outperforms both classical and deep learning-based clustering methods, offering a practical and theoretically grounded solution for unsupervised analysis of tabular data.

\paragraph{Limitations.}
Since \our{} is technically based on the TabPFN architecture, it inherits some of its drawbacks: a maximum number of input features and samples. However, the recently introduced TabPFN v2 \citep{hollmann2025accurate} showed that tabular transformers can process larger datasets with a negligible increase in computational time. Moreover, our experimental results demonstrate that even if the dimension of input data exceeds the fixed value of 30, applying PCA does not significantly hurt the clustering performance. 

The final performance of \our{} heavily relies on the synthetic data used in pre-training. While we release a basic version of \our{}, which encodes certain assumptions about data clusters, one can adjust the pre-training stage using different datasets. Finally, \our{} does not cluster data itself, but constructs a convenient embedding space in which clustering can be performed using basic algorithms.

\begin{ack}
We are grateful to the Reviewers for their effort and insightful comments on the paper.

This research is part of the project No. \textbf{2022/45/P/ST6/02969} co-funded by the National
Science Centre and the European Union Framework Programme for Research and
Innovation Horizon 2020 under the Marie Skłodowska-Curie grant agreement No.
945339. 
For the purpose of Open Access, the authors have applied a CC-BY public copyright licence to any Author Accepted Manuscript (AAM) version arising from this submission. 
\\
\includegraphics[width=1cm]{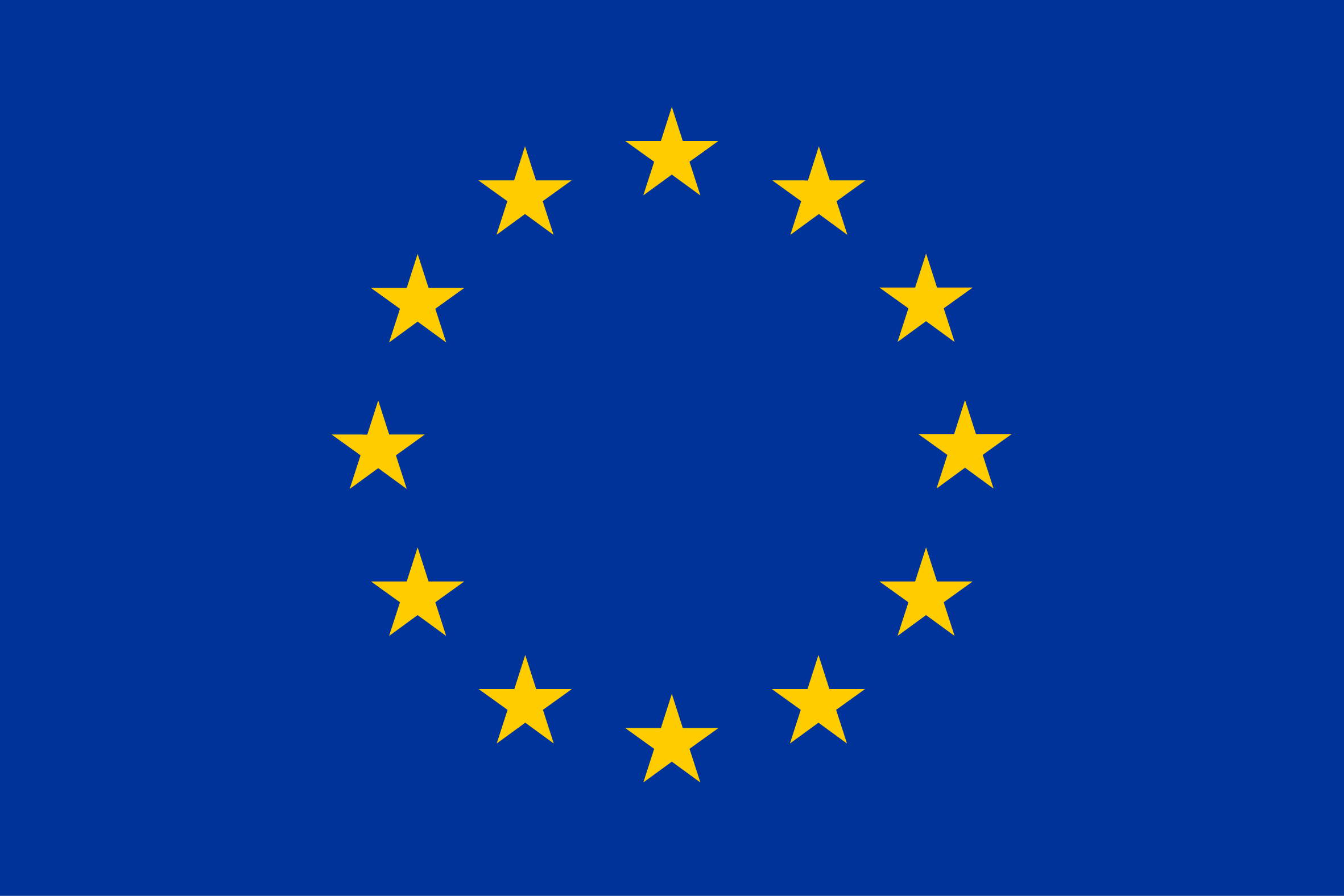} \includegraphics[width=1.9cm]{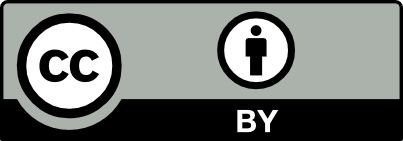}
\\ 
The research of P. Marszałek and M. Śmieja was supported by the National Science Centre (Poland), grant no. \textbf{2023/50/E/ST6/00169}. The research of Witold Wydmański was supported by the Ministry of Science grant no. \textbf{PN/01/0195/2022} and NCN Sonata BIS grant number \textbf{2020/38/E/NZ2/00598}. The research of J. Tabor was supported by the National Science Centre (Poland), grant no. \textbf{2023/49/B/ST6/01137}. Some experiments were performed on servers purchased with funds from
the flagship project entitled “Artificial Intelligence Computing Center Core Facility” from the DigiWorld Priority
Research Area within the Excellence Initiative – Research
University program at Jagiellonian University in Krakow. We also gratefully acknowledge Polish high-performance computing infrastructure PLGrid (HPC Center: ACK Cyfronet AGH) for providing computer facilities and support within computational grant no. \textbf{PLG/2024/017893}.

\end{ack}

\bibliographystyle{abbrv}
\bibliography{references}

\begin{thebibliography}{10}

\bibitem{scarf}
D.~Bahri, H.~Jiang, Y.~Tay, and D.~Metzler.
\newblock Scarf: Self-supervised contrastive learning using random feature
  corruption.
\newblock In {\em International Conference on Learning Representations}, 2022.

\bibitem{behrmann2019invertibleresidualnetworks}
J.~Behrmann, W.~Grathwohl, R.~T. Chen, D.~Duvenaud, and J.-H. Jacobsen.
\newblock Invertible residual networks.
\newblock In {\em {International Conference on Machine Learning}}, pages
  573--582. PMLR, 2019.

\bibitem{OpenML2025}
B.~Bischl, G.~Casalicchio, T.~Das, M.~Feurer, S.~Fischer, P.~Gijsbers,
  S.~Mukherjee, A.~C. Müller, L.~Németh, L.~Oala, L.~Purucker, S.~Ravi, J.~N.
  van Rijn, P.~Singh, J.~Vanschoren, J.~van~der Velde, and M.~Wever.
\newblock Openml: Insights from 10 years and more than a thousand papers.
\newblock {\em Patterns}, 6(7):101317, 2025.

\bibitem{brown2020languagemodelsfewshotlearners}
T.~Brown, B.~Mann, N.~Ryder, M.~Subbiah, J.~D. Kaplan, P.~Dhariwal,
  A.~Neelakantan, P.~Shyam, G.~Sastry, A.~Askell, et~al.
\newblock Language models are few-shot learners.
\newblock In {\em Advances in Neural Information Processing Systems},
  volume~33, pages 1877--1901, 2020.

\bibitem{cai2023semantic}
S.~Cai, L.~Qiu, X.~Chen, Q.~Zhang, and L.~Chen.
\newblock Semantic-enhanced image clustering.
\newblock In {\em Proceedings of the AAAI Conference on Artificial
  Intelligence}, volume~37, pages 6869--6878, 2023.

\bibitem{chen2020simple}
T.~Chen, S.~Kornblith, M.~Norouzi, and G.~Hinton.
\newblock A simple framework for contrastive learning of visual
  representations.
\newblock In {\em {International Conference on Machine Learning}}, pages
  1597--1607. PMLR, 2020.

\bibitem{dempster1977maximum}
A.~P. Dempster, N.~M. Laird, and D.~B. Rubin.
\newblock {Maximum likelihood from incomplete data via the EM algorithm}.
\newblock {\em Journal of the Royal Statistical Society: series B
  (methodological)}, 39(1):1--22, 1977.

\bibitem{8970928}
L.~Duan, C.~Aggarwal, S.~Ma, and S.~Sathe.
\newblock Improving spectral clustering with deep embedding and cluster
  estimation.
\newblock In {\em 2019 IEEE International Conference on Data Mining (ICDM)},
  pages 170--179, 2019.

\bibitem{garg2023transformerslearnincontextcase}
S.~Garg, D.~Tsipras, P.~Liang, and G.~Valiant.
\newblock What can transformers learn in-context? a case study of simple
  function classes.
\newblock In {\em Proceedings of the 36th International Conference on Neural
  Information Processing Systems}, NIPS '22, 2022.

\bibitem{dizaji2017deepclusteringjointconvolutional}
K.~Ghasedi~Dizaji, A.~Herandi, C.~Deng, W.~Cai, and H.~Huang.
\newblock Deep clustering via joint convolutional autoencoder embedding and
  relative entropy minimization.
\newblock In {\em Proceedings of the IEEE International Conference on Computer
  Vision}, pages 5736--5745, 2017.

\bibitem{grill2020bootstrap}
J.-B. Grill, F.~Strub, F.~Altch{\'e}, C.~Tallec, P.~Richemond, E.~Buchatskaya,
  C.~Doersch, B.~Avila~Pires, Z.~Guo, M.~Gheshlaghi~Azar, et~al.
\newblock Bootstrap your own latent-a new approach to self-supervised learning.
\newblock In {\em Advances in Neural Information Processing Systems},
  volume~33, pages 21271--21284, 2020.

\bibitem{grinsztajn2022tree}
L.~Grinsztajn, E.~Oyallon, and G.~Varoquaux.
\newblock Why do tree-based models still outperform deep learning on typical
  tabular data?
\newblock In {\em Advances in Neural Information Processing Systems},
  volume~35, pages 507--520, 2022.

\bibitem{guo_improved_2017}
X.~Guo, L.~Gao, X.~Liu, and J.~Yin.
\newblock Improved {Deep} {Embedded} {Clustering} with {Local} {Structure}
  {Preservation}.
\newblock In {\em Proceedings of the {Twenty}-{Sixth} {International} {Joint}
  {Conference} on {Artificial} {Intelligence}, {IJCAI}-17}, pages 1753--1759,
  2017.

\bibitem{han-etal-2023-understanding}
X.~Han, D.~Simig, T.~Mihaylov, Y.~Tsvetkov, A.~Celikyilmaz, and T.~Wang.
\newblock Understanding in-context learning via supportive pretraining data.
\newblock In {\em Proceedings of the 61st Annual Meeting of the Association for
  Computational Linguistics (Volume 1: Long Papers)}, pages 12660--12673.
  Association for Computational Linguistics, 2023.

\bibitem{hansen1982large}
L.~P. Hansen.
\newblock Large sample properties of generalized method of moments estimators.
\newblock {\em Econometrica: Journal of the Econometric Society}, pages
  1029--1054, 1982.

\bibitem{tabpfn}
N.~Hollmann, S.~M{\"u}ller, K.~Eggensperger, and F.~Hutter.
\newblock Tab{PFN}: A transformer that solves small tabular classification
  problems in a second.
\newblock In {\em The Eleventh International Conference on Learning
  Representations}, 2023.

\bibitem{hollmann2025accurate}
N.~Hollmann, S.~M{\"u}ller, L.~Purucker, A.~Krishnakumar, M.~K{\"o}rfer, S.~B.
  Hoo, R.~T. Schirrmeister, and F.~Hutter.
\newblock Accurate predictions on small data with a tabular foundation model.
\newblock {\em Nature}, 637(8045):319--326, 2025.

\bibitem{tabicl}
Q.~Jingang, D.~Holzm{\"u}ller, G.~Varoquaux, and M.~Le~Morvan.
\newblock {TabICL}: A tabular foundation model for in-context learning on large
  data.
\newblock In {\em Forty-second International Conference on Machine Learning},
  2025.

\bibitem{kuhn1955hungarian}
H.~W. Kuhn.
\newblock The {Hungarian} method for the assignment problem.
\newblock {\em Naval Research Logistics Quarterly}, 2(1-2):83--97, 1955.

\bibitem{kuhn1956variants}
H.~W. Kuhn.
\newblock Variants of the {Hungarian} method for assignment problems.
\newblock {\em Naval Research Logistics Quarterly}, 3(4):253--258, 1956.

\bibitem{lee2023supervised}
J.~Lee, A.~Xie, A.~Pacchiano, Y.~Chandak, C.~Finn, O.~Nachum, and E.~Brunskill.
\newblock Supervised pretraining can learn in-context reinforcement learning.
\newblock In {\em Thirty-seventh Conference on Neural Information Processing
  Systems}, 2023.

\bibitem{li2024image}
Y.~Li, P.~Hu, D.~Peng, J.~Lv, J.~Fan, and X.~Peng.
\newblock Image clustering with external guidance.
\newblock In {\em {International Conference on Machine Learning}}, pages
  27890--27902. PMLR, 2024.

\bibitem{macqueen1967some}
J.~MacQueen.
\newblock Some methods for classification and analysis of multivariate
  observations.
\newblock In {\em Proceedings of the Fifth Berkeley Symposium on Mathematical
  Statistics and Probability, Volume 1: Statistics}, volume~5, pages 281--298.
  University of California press, 1967.

\bibitem{mcelfresh2023neural}
D.~McElfresh, S.~Khandagale, J.~Valverde, V.~Prasad~C, G.~Ramakrishnan,
  M.~Goldblum, and C.~White.
\newblock When do neural nets outperform boosted trees on tabular data?
\newblock In {\em Advances in Neural Information Processing Systems},
  volume~36, pages 76336--76369, 2023.

\bibitem{bayes_in_context}
S.~M{\"u}ller, N.~Hollmann, S.~P. Arango, J.~Grabocka, and F.~Hutter.
\newblock Transformers can do {Bayesian} inference.
\newblock In {\em International Conference on Learning Representations}, 2022.

\bibitem{murtagh2012algorithms}
F.~Murtagh and P.~Contreras.
\newblock Algorithms for hierarchical clustering: an overview.
\newblock {\em Wiley Interdisciplinary Reviews: Data Mining and Knowledge
  Discovery}, 2(1):86--97, 2012.

\bibitem{rabbani2025deep}
S.~B. Rabbani, I.~V. Medri, and M.~D. Samad.
\newblock Deep clustering of tabular data by weighted {Gaussian} distribution
  learning.
\newblock {\em Neurocomputing}, 623:129359, 2025.

\bibitem{9506051}
M.~Sadeghi and N.~Armanfard.
\newblock {IDECF}: Improved deep embedding clustering with deep fuzzy
  supervision.
\newblock In {\em 2021 IEEE International Conference on Image Processing
  (ICIP)}, pages 1009--1013, 2021.

\bibitem{sadeghi2024deepclusteringselfsupervisionusing}
M.~Sadeghi, S.~Soleimani, and N.~Armanfard.
\newblock Deep clustering with self-supervision using pairwise similarities.
\newblock {\em IEEE Access}, 2025.

\bibitem{idc}
J.~Svirsky and O.~Lindenbaum.
\newblock Interpretable deep clustering for tabular data.
\newblock In {\em Proceedings of the 41st International Conference on Machine
  Learning}, volume 235, pages 47314--47330. PMLR, 21--27 Jul 2024.

\bibitem{transformer}
A.~Vaswani, N.~Shazeer, N.~Parmar, J.~Uszkoreit, L.~Jones, A.~N. Gomez,
  {\L}.~Kaiser, and I.~Polosukhin.
\newblock Attention is all you need.
\newblock In {\em Advances in Neural Information Processing Systems},
  volume~30, 2017.

\bibitem{xie2016unsuperviseddeepembeddingclustering}
J.~Xie, R.~Girshick, and A.~Farhadi.
\newblock Unsupervised deep embedding for clustering analysis.
\newblock In {\em {International Conference on Machine Learning}}, pages
  478--487. PMLR, 2016.

\bibitem{yang2017towards}
B.~Yang, X.~Fu, N.~D. Sidiropoulos, and M.~Hong.
\newblock Towards k-means-friendly spaces: Simultaneous deep learning and
  clustering.
\newblock In {\em {International Conference on Machine Learning}}, pages
  3861--3870. PMLR, 2017.

\bibitem{yoon2020vime}
J.~Yoon, Y.~Zhang, J.~Jordon, and M.~Van~der Schaar.
\newblock Vime: Extending the success of self-and semi-supervised learning to
  tabular domain.
\newblock In {\em Advances in Neural Information Processing Systems},
  volume~33, pages 11033--11043, 2020.

\bibitem{zhao2024comparison}
Z.~Zhao, L.~Alzubaidi, J.~Zhang, Y.~Duan, and Y.~Gu.
\newblock A comparison review of transfer learning and self-supervised
  learning: Definitions, applications, advantages and limitations.
\newblock {\em Expert Systems with Applications}, 242:122807, 2024.

\bibitem{zhu2023xtab}
B.~Zhu, X.~Shi, N.~Erickson, M.~Li, G.~Karypis, and M.~Shoaran.
\newblock {XTab}: cross-table pretraining for tabular transformers.
\newblock In {\em Proceedings of the 40th International Conference on Machine
  Learning}, pages 43181--43204, 2023.

\bibitem{znalezniak2023contrastive}
M.~Znalezniak, P.~Rola, P.~Kaszuba, J.~Tabor, and M.~{\'S}mieja.
\newblock Contrastive hierarchical clustering.
\newblock In {\em Joint European Conference on Machine Learning and Knowledge
  Discovery in Databases}, pages 627--643. Springer, 2023.

\end{thebibliography}

\clearpage
\appendix

\section*{\our{}: Zero-shot Embeddings for Unsupervised Separation of Tabular Data -- supplementary material}

\section{Background on Prior-Data Fitted Networks}
\label{appendix:pfn}

The recently introduced framework of \emph{Bayesian inference through transformers} demonstrates that neural networks pre-trained on synthetic datasets implicitly approximate Bayesian inference without explicitly computing posterior distributions \cite{bayes_in_context, tabpfn}. 
The pre-trained transformers are known as \emph{Prior-Data Fitted Networks} (PFNs). 
The pre-training involves synthetic prior fitting, wherein a transformer network is trained offline on numerous datasets generated from a predefined prior distribution over tasks. 
This pre-training procedure allows the transformer to implicitly encode a Bayesian posterior predictive distribution by optimizing a cross-entropy loss between its predictions and the synthetic data labels.  

Formally, pre-training is carried out as follows: given a prior distribution $p(\D)$ over (synthetic) datasets $\D$, a PFN network parameterized by $\theta$ is trained to minimize the negative log-likelihood of predicting held-out labels within datasets sampled from this prior. The associated loss function, known as the Prior-Data Negative Log-Likelihood (Prior-Data NLL), is explicitly defined as:
$$
\Loss_{\textit{PFN}}(\theta)  =  \mathbb{E}_{\D_\c\cup \{(x,y)\}\sim p(\D)}[-\log q_{\theta}(y|x,\D_\c)],
$$
where each dataset $D_\c$ and data point $(x,y)$ are sampled from the predefined prior distribution $p(D)$. As shown formally in \cite{bayes_in_context}, minimizing this loss is mathematically equivalent to minimizing the expected cross-entropy between the predictive distribution $q_{\theta}(y|x,D_\c)$ and the true posterior predictive distribution (PPD) derived from the prior. Specifically, minimizing the Prior-Data NLL is equivalent to minimizing the expected Kullback–Leibler divergence between the network's predictive distribution and the true PPD $p(y|x,D_\c)$:
$$
\Loss_{\textit{PFN}}(\theta)  =  \mathbb{E}_{\D_\c,x\sim p(D)}[H(p(\cdot|x,\D), q_{\theta}(\cdot|x,D_\c))],
$$
where $H$ denotes the cross-entropy. Thus, optimality of the predictive distribution $q_{\theta}$ implies matching it exactly to the true Bayesian posterior predictive distribution, provided that the parametric family of distributions defined by the transformer is sufficiently expressive \cite{bayes_in_context}.  

Within this in-context learning as Bayesian inference framework, transformers approximate Bayesian averaging implicitly. Given a training dataset $\D_\c  =  \{(x_{i},y_{i})\}_{i = 1}^{n}$, a new input $x$, and a prior over hypotheses $\phi$, the Bayesian posterior predictive distribution is formally expressed as:
$$
p(y|x,\D_\c)  =  \int_{\Phi} p(y|x,\phi)p(\D_\c|\phi)p(\phi)d\phi,
$$
integrating over all hypotheses $\phi \in \Phi$. PFNs approximate this integral implicitly through pre-training on prior-generated datasets, yielding a distribution $q_{\theta}(y|x,\D_\c)$ directly from forward propagation of the network conditioned on dataset $\D_\c$ \cite{bayes_in_context, tabpfn}.

\section{Details of synthetic data generation process}

\subsection{Theory for synthetic-to-real data generalization} \label{app:theory}

\begin{theorem}[Univeral Approximation Theorem for Mixture Distributions]
\label{theorem}
Let $P = \sum_{i=1}^k \pi_i P_i$ be a mixture of distributions on $\mathbb{R}^d$, where each $P_i$ is a probability distribution, and $\pi_i > 0$, $\sum_{i=1}^k \pi_i = 1$. 

Then there exists a mixture of Gaussians $Q = \sum_{i=1}^k \pi_i \mathcal{N}(\mu_i, \Sigma_i)$ and a neural network $F$ such that we can approximate $P$ with arbitrarily small error by the pushforward measure $F_{\#} Q$. Additionally, we can select $F$ so that $F_\# N(\mu_i,\Sigma_i)$ approximates $P_i$ with arbitrarily small error.
\end{theorem}
\begin{proof}
We proceed in several steps.

\textbf{Step 1: Approximation of individual components.} \\
    Making use of the universal approximation theorem, for each component $P_i$ there exists a neural network $F_i$ which transforms an arbitrary Gaussian $Q_i = \mathcal{N}(\mu_i,\Sigma_i)$ into $P_i$, i.e. $F_{i \#} Q_i \approx P_i$.

\textbf{Step 2: Constructing a unified neural network.} \\
We now define a single neural network $F$ that can represent all the $F_i$ networks. Define $F : \mathbb{R}^d \times \{1, \dots, k\} \to \mathbb{R}^d$ such that $F(z, i) \approx F_i(z)$. This can be implemented by encoding the index $i$ as a one-hot vector or embedding and concatenating it to $z$, enabling $F$ to learn the behavior of each $F_i$ conditioned on the index $i$. The network $F$ can be extended to the domain $\mathbb{R}^d \times \mathbb{R}$ in an arbitrary way.

\textbf{Step 3: Approximation of the full mixture.} \\
Since each $F_{\#} \mathcal{N}(\mu_i, \Sigma_i)$ approximates $P_i$, the mixture $
F_{\#} Q = \sum_{i=1}^k \pi_i F_{\#} \mathcal{N}(\mu_i, \Sigma_i)$ approximates $P = \sum_{i=1}^k \pi_i P_i$.

Thus, we have constructed a mixture of Gaussians $Q$ and a neural network $F$ such that $F_{\#} Q$ approximates $P$, and each component $F_{\#} \mathcal{N}(\mu_i, \Sigma_i)$ approximates $P_i$ arbitrarily well. This completes the proof.
\end{proof}

\subsection{Insights into data-generating priors}
\label{sec:gen}

We use three types of probabilistic procedures to generate the prior data:
\begin{enumerate}
    \item \textbf{Gaussian}. The goal is to create a continuous data type where each cluster follows a multivariate Gaussian distribution with carefully designed means and covariance matrices. This construction process is incremental. Starting with a mixture containing $k$ components, the addition of a new cluster involves initially placing it at position 0 and then shifting it in a randomly chosen direction. The covariance structure for each Gaussian is generated through eigendecomposition, with eigenvalues sampled from a predefined range of $[0.005, 0.05]$ to control the shape and orientation of the clusters. Additionally, to guarantee the presence of both full-rank Gaussians in higher dimensions and degenerate ones in lower dimensions, extra conditions are introduced to narrow the range of eigenvalues. These constraints are activated with probabilities of 0.25 and 0.2, respectively. To ensure adequate separation between clusters and to prevent trivial overlap, we apply a Wasserstein-2 distance constraint, requiring that the means $\mu_i$ and $\mu_j$ of any two clusters, along with their corresponding covariance matrices $\Sigma_i$ and $\Sigma_j$, are separated by at least a threshold $T$, as follows:
    \begin{equation}
    \label{eq:gaussian_constraint}
     \|\mu_i-\mu_j\|^2_2 + tr(\Sigma_i + \Sigma_j - 2(\Sigma_i^{\frac{1}{2}}\Sigma_j\Sigma_i^{\frac{1}{2}})^{\frac{1}{2}}) \geq T
    \end{equation}
    The exact minimum distance value $T$ varies between 0.5 and 1.0 and is randomly selected for each component independently in order to promote data diversity. This entire approach allows us to model a variety of cluster geometries, from spherical to highly elongated elliptical shapes.
    \item \textbf{Categorical}. To enrich the generated data, we incorporate categorical features alongside continuous ones by sampling from categorical distributions that are biased toward certain categories for each cluster. To produce these varied categorical probability patterns, we use the Dirichlet distribution. The probability of including categorical features is controlled by a $categorical\_chance$ parameter, set to 0.3. Up to 3 categorical variables may be added, each having between 2 and $max\_categories$ possible values, defined as 5. The resulting categorical variables are then converted to one-hot encoding and combined with the continuous features, producing mixed-type datasets that more closely resemble real-world tabular data.
    \item \textbf{NN transformed}. To create more complex, non-linearly separable cluster structures, we apply transformations to the numerical features using random neural networks with 3 to 6 layers. These transformations map the original data through several non-linear operations while preserving cluster identity information, producing datasets with more challenging decision boundaries. Following the approach of Invertible Residual Networks~\cite{behrmann2019invertibleresidualnetworks}, we constrain the spectral norm of each transformation layer to be less than 1. In addition, standardization is applied between residual layers to ensure more stable transformations. To help preserve cluster separability, we append one-hot vectors indicating the component identity to the numerical variables before passing them through the random neural network. After the transformation, these extra dimensions are removed using the PCA algorithm, restoring the data to its original dimensionality. This comprehensive approach enables the simulation of intricate, non-linear data manifold structures often presented in real-world clustering problems without introducing degenerate configurations.
\end{enumerate}

All features undergo standardization and scaling to ensure numerical stability during training. Continuous features are normalized to the range $[-1, 1]$, while ensuring that the relative separation between clusters is preserved.

By training on this diverse collection of synthetic datasets-ranging from well-separated Gaussian clusters to complex, transformed manifolds with mixed feature types \our{} learns to identify meaningful cluster structures across a wide spectrum of data distributions. This enables it to adapt to previously unseen datasets at inference time without additional training.

\section{Pre-training plots}
\label{appendix:pre-training graphs}

\begin{figure}[!ht]
    \centering
    \includegraphics[width=0.49\textwidth]{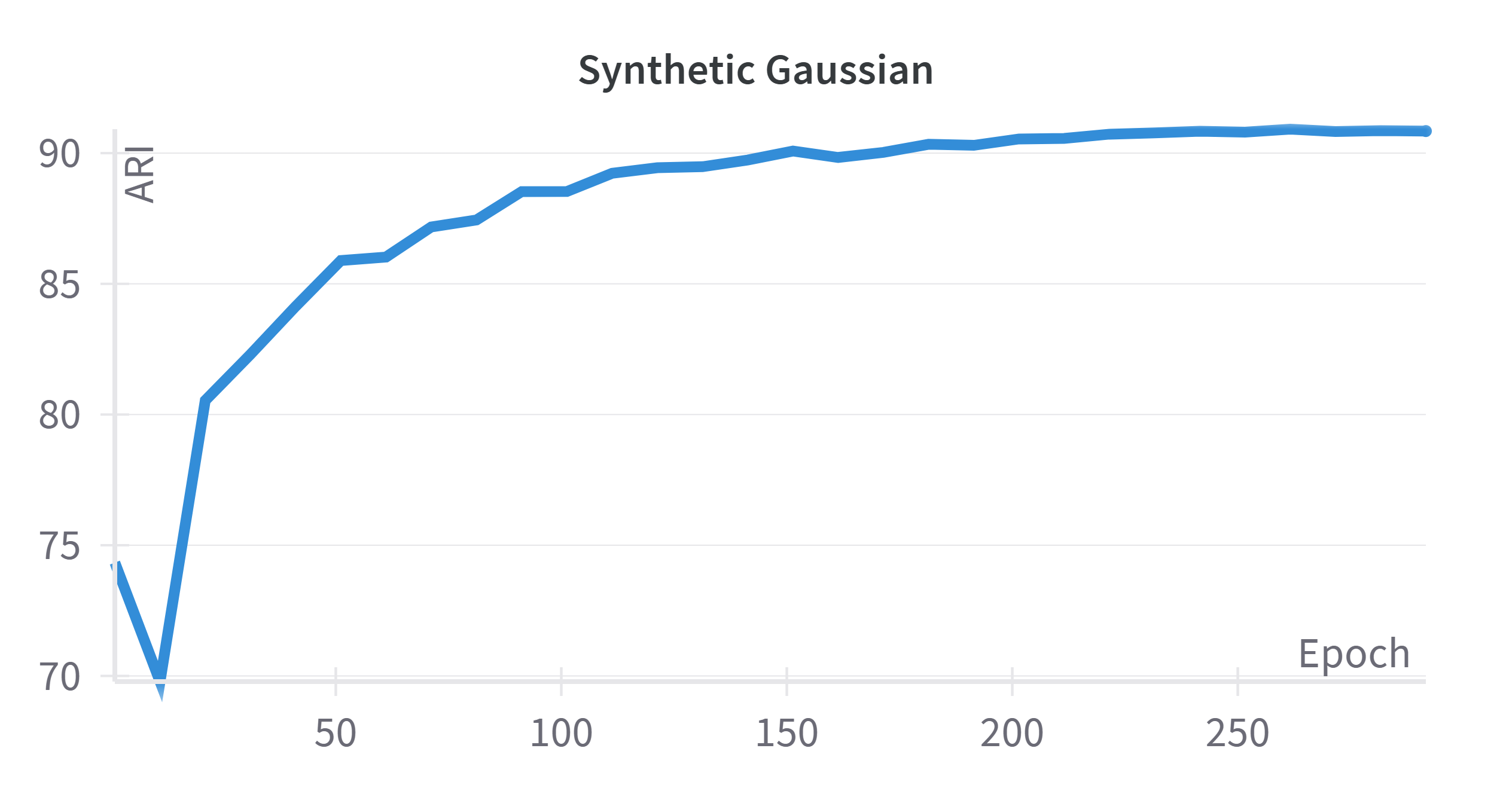}
    \includegraphics[width=0.49\textwidth]{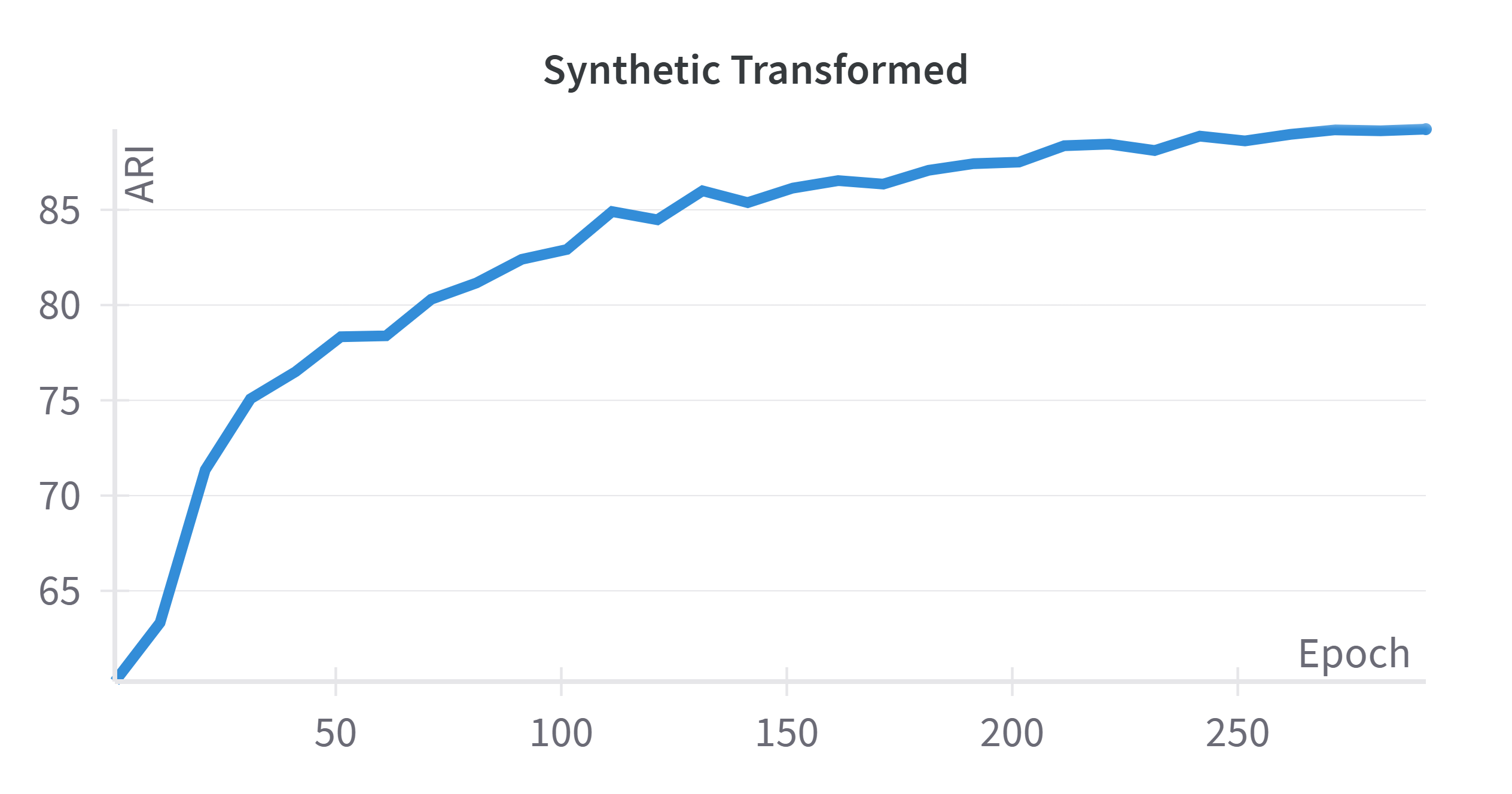}
    \caption{Visualization of pre-training process}
    \label{fig:pre-training_plot}
\end{figure}

Figure~\ref{fig:pre-training_plot} presents two plots of average ARI over 200 synthetic validation datasets throughout 300 pre-training epochs. The first plot corresponds to the Gaussian-categorical datasets referred to as Synthetic Gaussian, while the second illustrates their NN transformed variants, named Synthetic Transformed. The data generation procedures are described in more detail in Appendix~\ref{sec:gen}. Both plots illustrate that the quality of the \our{} representation improves over time, indicating that the model is learning new patterns, as evidenced by the increasing ARI during the pre-training process.

\section{Statistics of datasets used in experimental study}
\label{appendix:statistics}

\begin{table}[ht!]
    \caption{Real/OpenML datasets statistics.}
    \label{tab:statistics_real}
    \centering
    \begin{tabular}{lcccccc}
        \toprule
        ID & \# Instances & \# Numerical & \# Categorical & Dimension & \# Categories & \# Classes \\
         &  & features & features &  &  (one-hots) &  \\
        \midrule
        14 & 2000 & 76 & 0 & 76 & 0 & 10 \\
        15 & 699 & 9 & 0 & 9 & 0 & 2 \\
        16 & 2000 & 64 & 0 & 64 & 0 & 10 \\
        18 & 2000 & 6 & 0 & 6 & 0 & 10 \\
        22 & 2000 & 47 & 0 & 47 & 0 & 10 \\
        35 & 366 & 1 & 33 & 130 & 129 & 6 \\
        51 & 294 & 6 & 7 & 25 & 19 & 2 \\
        53 & 270 & 13 & 0 & 13 & 0 & 2 \\
        56 & 435 & 0 & 16 & 32 & 32 & 2 \\
        61 & 150 & 4 & 0 & 4 & 0 & 3 \\
        187 & 178 & 13 & 0 & 13 & 0 & 3 \\
        377 & 600 & 60 & 0 & 60 & 0 & 6 \\
        458 & 841 & 70 & 0 & 70 & 0 & 4 \\
        481 & 209 & 7 & 1 & 14 & 7 & 2 \\
        694 & 310 & 8 & 0 & 8 & 0 & 9 \\
        721 & 200 & 10 & 0 & 10 & 0 & 2 \\
        733 & 209 & 6 & 0 & 6 & 0 & 2 \\
        745 & 159 & 14 & 1 & 20 & 6 & 2 \\
        756 & 159 & 15 & 0 & 15 & 0 & 2 \\
        796 & 209 & 6 & 1 & 36 & 30 & 2 \\
        820 & 235 & 12 & 0 & 12 & 0 & 2 \\
        840 & 205 & 17 & 8 & 68 & 51 & 2 \\
        854 & 158 & 5 & 2 & 14 & 9 & 2 \\
        1462 & 1372 & 4 & 0 & 4 & 0 & 2 \\
        1495 & 250 & 0 & 6 & 18 & 18 & 2 \\
        1499 & 210 & 7 & 0 & 7 & 0 & 3 \\
        1510 & 569 & 30 & 0 & 30 & 0 & 2 \\
        1523 & 310 & 6 & 0 & 6 & 0 & 3 \\
        4153 & 180 & 66 & 0 & 66 & 0 & 6 \\
        40496 & 500 & 7 & 0 & 7 & 0 & 10 \\
        40682 & 215 & 5 & 0 & 5 & 0 & 3 \\
        40705 & 959 & 42 & 2 & 44 & 2 & 2 \\
        42261 & 150 & 4 & 0 & 4 & 0 & 3 \\
        42585 & 344 & 4 & 2 & 10 & 6 & 3 \\
        \bottomrule
    \end{tabular}
\end{table}

\begin{table}[ht!]
    \caption{Synthetic Gaussian datasets statistics.}
    \label{tab:statistics_gaussian}
    \centering
    \begin{tabular}{lcccccc}
        \toprule
        ID & \# Instances & \# Numerical & \# Categorical & Dimension & \# Categories & \# Classes \\
         &  & features & features &  &  (one-hots) &  \\
        \midrule
        0 & 1337 & 16 & 0 & 16 & 0 & 8 \\
        1 & 1383 & 23 & 0 & 23 & 0 & 8 \\
        2 & 1421 & 8 & 2 & 14 & 6 & 7 \\
        3 & 992 & 9 & 1 & 12 & 3 & 9 \\
        4 & 1314 & 8 & 0 & 8 & 0 & 9 \\
        5 & 1497 & 16 & 2 & 23 & 7 & 8 \\
        6 & 1370 & 2 & 0 & 2 & 0 & 6 \\
        7 & 1646 & 4 & 3 & 16 & 12 & 8 \\
        8 & 1520 & 11 & 0 & 11 & 0 & 8 \\
        9 & 1537 & 18 & 0 & 18 & 0 & 5 \\
        10 & 825 & 26 & 0 & 26 & 0 & 2 \\
        11 & 1112 & 9 & 0 & 9 & 0 & 5 \\
        12 & 1093 & 15 & 0 & 15 & 0 & 8 \\
        13 & 742 & 6 & 0 & 6 & 0 & 3 \\
        14 & 1595 & 11 & 2 & 18 & 7 & 7 \\
        15 & 1417 & 14 & 0 & 14 & 0 & 6 \\
        16 & 1787 & 28 & 0 & 28 & 0 & 5 \\
        17 & 764 & 19 & 0 & 19 & 0 & 4 \\
        18 & 889 & 25 & 0 & 25 & 0 & 5 \\
        19 & 1660 & 28 & 0 & 28 & 0 & 9 \\
        \bottomrule
    \end{tabular}
\end{table}

\begin{table}[ht!]
    \caption{Synthetic Transformed datasets statistics.}
    \label{tab:statistics_transformed}
    \centering
    \begin{tabular}{lcccccc}
        \toprule
        ID & \# Instances & \# Numerical & \# Categorical & Dimension & \# Categories & \# Classes \\
         &  & features & features &  &  (one-hots) &  \\
        \midrule
        0 & 1337 & 16 & 0 & 16 & 0 & 8 \\
        1 & 1627 & 30 & 0 & 30 & 0 & 9 \\
        2 & 1631 & 11 & 0 & 11 & 0 & 7 \\
        3 & 1891 & 15 & 1 & 17 & 2 & 10 \\
        4 & 1142 & 3 & 0 & 3 & 0 & 4 \\
        5 & 1222 & 24 & 0 & 24 & 0 & 9 \\
        6 & 953 & 6 & 0 & 6 & 0 & 6 \\
        7 & 1508 & 9 & 0 & 9 & 0 & 10 \\
        8 & 840 & 7 & 3 & 15 & 8 & 5 \\
        9 & 1745 & 14 & 0 & 14 & 0 & 9 \\
        10 & 1618 & 23 & 0 & 23 & 0 & 6 \\
        11 & 1432 & 13 & 0 & 13 & 0 & 9 \\
        12 & 1860 & 9 & 0 & 9 & 0 & 9 \\
        13 & 563 & 10 & 0 & 10 & 0 & 2 \\
        14 & 1033 & 6 & 2 & 13 & 7 & 3 \\
        15 & 750 & 2 & 0 & 2 & 0 & 4 \\
        16 & 1451 & 14 & 2 & 20 & 6 & 10 \\
        17 & 679 & 30 & 0 & 30 & 0 & 2 \\
        18 & 859 & 22 & 1 & 25 & 3 & 2 \\
        19 & 1493 & 11 & 0 & 11 & 0 & 7 \\
        \bottomrule
    \end{tabular}
\end{table}

Tables \ref{tab:statistics_real}, \ref{tab:statistics_gaussian}, and \ref{tab:statistics_transformed} provide detailed information about the datasets used in the experimental analysis. Each table includes the number of set instances, the number of categorical and numerical features, the total dimensionality, and the overall number of categories, represented by the length of the one-hot encoded vectors. Additionally, the tables report the number of classes that each dataset contains. In Table \ref{tab:statistics_real}, the ID column corresponds to the OpenML ID, while in the remaining tables it functions solely as an index.

\FloatBarrier

\section{Baselines}
\label{appendix:baselines}
The evaluation of baseline models is based on the following libraries and GitHub repositories:
\begin{enumerate}
    \item \textbf{\href{https://scikit-learn.org/stable/}{scikit-learn}} - used for k-means and GMM,
    \item \textbf{\href{https://github.com/vlukiyanov/pt-dec}{https://github.com/vlukiyanov/pt-dec}} - implementation of the DEC,
    \item \textbf{\href{https://github.com/dawnranger/IDEC-pytorch}{https://github.com/dawnranger/IDEC-pytorch}} - source code for the IDEC,
    \item 
    \textbf{\href{https://github.com/jsvir/idc}{https://github.com/jsvir/idc}} - official implementation of the IDC method,
    \item \textbf{\href{https://github.com/mdsamad001/G-CEALS---Deep-Clustering-for-Tabular-Data}{https://github.com/mdsamad001/G-CEALS---Deep-Clustering-for-Tabular-Data}} - codebase for the GCEALS,
    \item 
    \textbf{\href{https://github.com/PriorLabs/tabpfn-extensions}{https://github.com/PriorLabs/tabpfn-extensions}} - a library that extends TabPFN functionality to a wide spectrum of machine learning tasks, including unsupervised ones,
    \item 
    \textbf{\href{https://github.com/clabrugere/pytorch-scarf}{https://github.com/clabrugere/pytorch-scarf}} - code repository for SCARF.
\end{enumerate}

To ensure fair comparison, hyperparameters are chosen to maximize the performance of each method with respect to their overall average rank across 5 random seeds. For this reason, all numerical features are preprocessed using a standard scaler prior to the training phase.

Most parameters of the k-means and GMM methods are left at their default values. Only the $n\_init$ option was increased to 100 for k-means and 50 for GMM in order to improve stability.

As mentioned in Section~\ref{sec:baselines}, a standard autoencoder with hidden layers $[500, 500, 2000]$ is used for each AE-based method. The resulting network is first pre-trained for 1000 epochs and then fine-tuned for up to 1000 additional epochs, separately for each dataset and model. Following the GCEALS evaluation procedure, multiple latent dimension sizes $[5, 10, 15, 20]$ were tested across all considered methods. The results indicate that the default latent dimension of 10 does not yield the best performance; instead, a dimension of 20 generally performs better. Changing other hyperparameters, including the learning rate, optimizer, and clustering loss weight $\gamma$, generally did not lead to improved results. Therefore, the remaining parameters were left at their default values as proposed by the authors of the respective codebases.

For TabPFN, we use the \textit{get\_embeddings\_per\_column} method from the \textit{TabPFNUnsupervisedModel} class, provided in the TabPFN Extensions repository\footnote{\url{https://github.com/PriorLabs/tabpfn-extensions}}, to extract per-column representations. These embeddings are then averaged across columns to obtain the final representation for each data point. For SCARF, we follow the procedure outlined in the \textit{example.ipynb} notebook from the original repository\footnote{\url{https://github.com/clabrugere/pytorch-scarf}}. In both cases, we cluster the learned representations using $k$-means, consistent with our approach in \our{}.

\section{Extended experimental results} \label{sec:exp_app}

This section presents extended tables corresponding to the experiments described in Section~\ref{sec:experiments}.

\subsection{How effective is \our{} for clustering?}
\label{appendix:extended_results_ari}

Tables~\ref{tab:extended_real_ari}, \ref{tab:extended_gauss_ari}, and \ref{tab:extended_transf_ari} contain the complete results for individual datasets from the experiments discussed in Section~\ref{sec:quantitative_results} in Tables~\ref{tab:comp_ari} and \ref{tab:comp_rank}, as presented in the main text. The reported values represent averages of ARI over 5 random seeds. In addition to the rows presenting outcomes for specific datasets, the tables include summary rows labeled Mean, Mean-Rank, Top-3, and Top-1. These represent, respectively: the average ARI, the average rank computed across all datasets, the number of times a given model appears in the top 3, and the total number of wins achieved by each method. It is worth noting that the Top-1 and Top-3 rows indicate clear wins and clear appearances in top-3 positions.

The overall conclusions align with those presented in the primary analysis. One noteworthy point is that \our{} achieves strong performance in terms of the Top-3 and Top-1 statistics, being outperformed in this regard only in Table~\ref{tab:extended_gauss_ari}, where k-means and GMM score better. Another interesting observation is that GMM struggles with categorical features, which significantly worsens its average scores in Tables~\ref{tab:extended_gauss_ari} and~\ref{tab:extended_transf_ari}.

\begin{table}[ht!]
\scriptsize
\caption{Comparison of clustering quality using ARI metric on OpenML datasets (higher score indicate better performance).}
\label{tab:extended_real_ari}
\centering
\begin{tabular}{lccccccccccc}
\toprule
ID & KM & GMM & AE-KM & AE-GMM & DEC & IDEC & IDC & G-CEALS & TabPFN & SCARF & \our \\
\midrule
14 & 38.74 & 45.90 & 40.94 & 45.36 & \underline{49.60} & 36.78 & 45.03 & 45.04 & 22.02 & 1.90 & \textbf{50.56} \\
15 & 82.85 & 71.38 & 61.44 & 71.74 & \underline{86.29} & \textbf{87.48} & 83.20 & 24.38 & 74.31 & 75.01 & 81.28 \\
16 & 55.62 & 64.75 & 61.66 & \underline{69.97} & 68.56 & 55.61 & 65.84 & 55.27 & 14.69 & 3.74 & \textbf{74.03} \\
18 & \textbf{54.86} & 46.88 & 51.45 & 51.47 & 50.00 & \underline{53.54} & 44.40 & 49.93 & 51.50 & 5.02 & 51.63 \\
22 & 35.48 & 47.39 & 31.09 & \underline{50.80} & 50.02 & 35.69 & 46.03 & 28.38 & 11.09 & 2.63 & \textbf{56.05} \\
35 & 70.68 & 70.68 & 70.22 & 70.48 & 68.63 & 64.58 & \underline{77.99} & 56.57 & 52.24 & 74.61 & \textbf{85.12} \\
51 & 28.35 & -2.54 & 37.82 & 34.13 & 38.13 & 34.58 & \underline{38.93} & 35.13 & 38.26 & -1.27 & \textbf{43.66} \\
53 & \textbf{45.21} & 5.17 & 39.14 & 40.62 & \underline{43.50} & 41.91 & 33.03 & 21.32 & 26.61 & 24.24 & 35.76 \\
56 & 57.79 & 58.49 & 57.10 & 58.31 & 56.97 & 55.20 & 59.25 & 55.66 & 58.42 & \underline{61.26} & \textbf{66.41} \\
61 & 62.01 & \textbf{90.39} & 59.79 & 60.21 & 57.84 & 60.14 & 58.30 & 51.37 & 77.20 & 51.19 & \underline{85.15} \\
187 & \underline{89.75} & \textbf{93.09} & 82.87 & 84.86 & 85.02 & 85.10 & 80.04 & 55.67 & 37.55 & 40.05 & 88.19 \\
377 & 56.70 & 58.47 & 62.06 & 59.52 & \underline{63.03} & 59.01 & 62.39 & 59.79 & \textbf{66.89} & 14.45 & 55.30 \\
458 & 95.08 & \underline{96.97} & 64.33 & 75.70 & 94.58 & 63.29 & 86.77 & 68.31 & 61.41 & 2.87 & \textbf{99.19} \\
481 & \textbf{58.51} & 2.77 & 36.84 & 48.03 & \underline{57.28} & 52.93 & 43.17 & 19.67 & 2.77 & 1.79 & 8.68 \\
694 & 35.91 & \textbf{43.30} & 35.26 & \underline{36.56} & 26.30 & 36.02 & 33.47 & 30.76 & 27.68 & 14.09 & 33.38 \\
721 & \textbf{43.28} & \underline{43.28} & 13.12 & 25.12 & 17.75 & 28.77 & 12.74 & -0.03 & 23.63 & 6.35 & 43.28 \\
733 & 50.97 & 46.29 & 38.39 & 55.11 & 67.47 & \underline{73.51} & 68.88 & 26.26 & -8.35 & 49.41 & \textbf{74.97} \\
745 & 57.63 & 55.44 & \textbf{78.35} & \underline{75.52} & 46.37 & 50.94 & 59.29 & 57.81 & 1.26 & 16.75 & 73.85 \\
756 & 55.70 & 4.46 & \textbf{72.81} & 53.40 & 58.60 & \underline{65.58} & 53.32 & 44.14 & 10.60 & 64.82 & 50.31 \\
796 & 49.62 & 2.26 & \underline{76.54} & 52.19 & 50.07 & 55.92 & \textbf{79.59} & 43.40 & 0.88 & 58.25 & 13.99 \\
820 & 50.16 & 36.77 & 44.29 & 49.24 & \textbf{51.21} & 49.30 & 44.59 & 24.51 & \underline{50.26} & 13.04 & 28.87 \\
840 & 39.05 & 48.37 & \textbf{53.60} & 46.72 & \underline{49.14} & 39.16 & 19.59 & 27.95 & 8.89 & 32.13 & 27.66 \\
854 & \textbf{76.14} & -0.21 & 62.33 & 60.83 & 35.05 & 70.52 & 64.81 & 72.25 & -0.54 & 6.48 & \underline{76.14} \\
1462 & 1.32 & 0.31 & 10.03 & \underline{10.17} & 1.00 & 3.80 & 8.83 & 8.48 & 3.31 & -0.03 & \textbf{92.28} \\
1495 & \underline{96.81} & \textbf{98.40} & 72.74 & 82.80 & 96.81 & 91.28 & 83.96 & 77.51 & 47.11 & 5.10 & 7.35 \\
1499 & 77.33 & 62.99 & 73.68 & 75.28 & 69.99 & \underline{78.48} & 67.82 & 29.60 & 5.82 & 24.93 & \textbf{82.40} \\
1510 & 67.07 & \textbf{78.02} & 63.73 & 59.77 & 71.32 & 69.84 & 70.55 & 69.47 & 54.59 & 63.57 & \underline{74.26} \\
1523 & 21.16 & \textbf{41.21} & 25.48 & 24.57 & 28.16 & 23.05 & \underline{31.35} & 18.50 & 2.45 & 21.04 & 19.60 \\
4153 & 55.78 & 58.03 & \underline{59.15} & 46.23 & 56.41 & 58.44 & 50.44 & 39.43 & 27.80 & 36.21 & \textbf{62.29} \\
40496 & \textbf{54.02} & 34.79 & 42.35 & 37.51 & \underline{49.74} & 42.35 & 27.41 & 37.33 & 31.80 & 39.78 & 32.25 \\
40682 & 58.32 & \underline{86.29} & 57.89 & 59.22 & 57.39 & 86.05 & 17.95 & 40.58 & \textbf{86.94} & 17.86 & 53.15 \\
40705 & 43.49 & 37.92 & 0.73 & 30.63 & \textbf{47.90} & \underline{46.28} & 31.38 & 10.71 & -4.40 & 32.90 & 45.45 \\
42261 & 62.01 & \textbf{90.39} & 60.13 & 58.15 & 59.78 & 56.78 & 55.10 & 49.09 & 77.20 & 52.53 & \underline{85.15} \\
42585 & 60.84 & 30.52 & 51.26 & 60.71 & \underline{91.79} & 43.40 & 72.10 & 38.30 & 23.04 & 3.50 & \textbf{95.05} \\
\midrule
Mean & 55.54 & 48.49 & 51.43 & 53.56 & \underline{55.93} & 54.57 & 52.28 & 40.37 & 31.32 & 26.95 & \textbf{57.43} \\
Mean-Rank & \underline{4.69} & 5.65 & 5.72 & 5.24 & \underline{4.69} & 5.01 & 5.62 & 8.18 & 8.22 & 8.85 & \textbf{4.13} \\
Top-3 & 11 & 12 & 6 & 5 & \underline{15} & 11 & 9 & 1 & 6 & 4 & \textbf{21} \\
Top-1 & 4 & \underline{7} & 3 & 0 & 2 & 1 & 1 & 0 & 2 & 0 & \textbf{12} \\
\bottomrule
\end{tabular}
\end{table}

\begin{table}[ht!]
\scriptsize
\caption{Evaluation of clustering quality with the ARI metric on Synthetic Gaussian datasets (higher scores reflect better performance).}
\label{tab:extended_gauss_ari}
\centering
\begin{tabular}{lccccccccccc}
\toprule
ID & KM & GMM & AE-KM & AE-GMM & DEC & IDEC & IDC & G-CEALS & TabPFN & SCARF & \our \\
\midrule
0 & 86.16 & 82.13 & 83.44 & 79.66 & \underline{87.06} & 77.26 & 60.64 & 58.57 & 65.42 & 4.06 & \textbf{87.65} \\
1 & \textbf{88.66} & \underline{87.36} & 81.13 & 79.61 & 85.84 & 75.62 & 44.44 & 54.10 & 40.26 & 4.21 & 72.04 \\
2 & 81.30 & 26.46 & 71.57 & 78.04 & \textbf{86.65} & 78.91 & 53.53 & 48.59 & 20.42 & 11.44 & \underline{82.96} \\
3 & \underline{95.02} & 36.77 & 92.03 & 76.44 & 92.74 & 89.81 & 65.03 & 62.67 & 50.57 & 22.29 & \textbf{95.76} \\
4 & 90.64 & \textbf{92.97} & 82.65 & 77.80 & 91.61 & 84.51 & 77.56 & 69.70 & 61.65 & 8.80 & \underline{92.85} \\
5 & 89.83 & 25.85 & 86.52 & 69.06 & \underline{90.01} & 88.31 & 70.38 & 67.70 & 20.51 & 10.31 & \textbf{90.72} \\
6 & 96.21 & \textbf{98.07} & 85.99 & 94.26 & 94.24 & 92.39 & 80.18 & 73.15 & 86.50 & 6.51 & \underline{97.89} \\
7 & \underline{92.59} & 30.52 & 81.73 & 79.06 & 91.35 & 89.80 & 46.44 & 58.10 & 26.62 & 10.14 & \textbf{93.60} \\
8 & \underline{88.29} & \textbf{89.62} & 81.15 & 76.53 & 87.56 & 78.37 & 72.72 & 50.30 & 68.54 & 13.06 & 84.00 \\
9 & 93.36 & \textbf{98.54} & 68.58 & 85.42 & 91.14 & 85.29 & 76.64 & 56.89 & 65.02 & 11.37 & \underline{94.55} \\
10 & 79.26 & \textbf{93.32} & 72.75 & \underline{90.90} & 72.54 & 73.51 & 77.64 & 48.47 & 89.15 & -0.04 & 87.32 \\
11 & 93.60 & \textbf{98.48} & 90.77 & 89.11 & 95.03 & 92.19 & 74.56 & 64.24 & 68.28 & 2.11 & \underline{97.45} \\
12 & \textbf{86.82} & 77.45 & 73.64 & 71.40 & \underline{86.04} & 77.93 & 60.55 & 58.45 & 55.27 & 19.11 & 86.04 \\
13 & \underline{96.77} & \textbf{98.07} & 94.05 & 90.15 & 96.11 & 85.66 & 86.14 & 78.20 & 80.24 & 5.29 & 96.17 \\
14 & 90.73 & 43.20 & 79.65 & 69.09 & \underline{92.68} & 77.93 & 72.80 & 67.84 & 12.93 & 8.16 & \textbf{94.60} \\
15 & 93.58 & \underline{96.23} & 85.74 & 89.30 & 94.43 & 89.41 & 72.54 & 72.88 & 91.02 & 3.24 & \textbf{96.68} \\
16 & \textbf{94.70} & 92.45 & 90.98 & 89.41 & \underline{94.53} & 88.14 & 77.56 & 85.86 & 59.58 & 4.85 & 87.19 \\
17 & 85.26 & \textbf{87.10} & 62.23 & 79.52 & 84.08 & 75.00 & 50.45 & 42.55 & 50.06 & 7.69 & \underline{86.29} \\
18 & \underline{88.77} & \textbf{94.83} & 84.47 & 84.74 & 87.39 & 83.46 & 61.36 & 67.96 & 74.90 & 7.69 & 83.01 \\
19 & \underline{86.54} & \textbf{89.11} & 76.18 & 78.46 & 86.06 & 67.99 & 47.46 & 70.50 & 32.39 & 6.02 & 73.85 \\
\midrule
Mean & \textbf{89.90} & 76.93 & 81.26 & 81.40 & \underline{89.35} & 82.57 & 66.43 & 62.84 & 55.97 & 8.32 & 89.03 \\
Mean-Rank & \textbf{2.65} & 3.65 & 5.65 & 5.60 & 3.23 & 5.70 & 7.95 & 8.90 & 8.75 & 11.00 & \underline{2.92} \\
Top-3 & \textbf{16} & 13 & 0 & 1 & \underline{15} & 0 & 0 & 0 & 1 & 0 & 14 \\
Top-1 & 3 & \textbf{10} & 0 & 0 & 1 & 0 & 0 & 0 & 0 & 0 & \underline{6} \\
\bottomrule
\end{tabular}
\end{table}

\begin{table}[ht!]
\scriptsize
\caption{Assessment of clustering quality based on the ARI metric on Synthetic Transformed datasets (higher score indicate better performance).}
\label{tab:extended_transf_ari}
\centering
\begin{tabular}{lccccccccccc}
\toprule
ID & KM & GMM & AE-KM & AE-GMM & DEC & IDEC & IDC & G-CEALS & TabPFN & SCARF & \our \\
\midrule
0 & 60.56 & \underline{79.33} & 46.31 & 68.98 & 76.93 & 46.35 & 56.67 & 29.42 & 6.36 & 0.94 & \textbf{87.42} \\
1 & \underline{93.30} & \textbf{96.55} & 76.57 & 83.99 & 91.29 & 70.91 & 66.59 & 73.09 & 16.59 & 1.22 & 89.40 \\
2 & 89.76 & \textbf{96.31} & 57.79 & 65.99 & 94.42 & 66.34 & 72.25 & 48.34 & 10.00 & 1.10 & \underline{95.88} \\
3 & 70.16 & 50.38 & 56.23 & 76.06 & \textbf{83.06} & 60.09 & 65.13 & 54.54 & 15.36 & 1.99 & \underline{80.98} \\
4 & 44.32 & \textbf{80.63} & 47.60 & 44.77 & 68.39 & 50.12 & 36.00 & 49.37 & 19.14 & 1.48 & \underline{73.66} \\
5 & \underline{91.49} & \textbf{96.51} & 74.48 & 89.11 & 86.46 & 77.84 & 60.00 & 66.47 & 15.12 & 1.25 & 81.63 \\
6 & 60.98 & \underline{81.63} & 61.97 & 76.29 & 75.58 & 61.89 & 69.43 & 69.84 & 18.83 & 2.11 & \textbf{87.02} \\
7 & 76.80 & 83.98 & 71.37 & 77.69 & \underline{86.56} & 77.49 & 76.75 & 56.51 & 18.94 & 3.20 & \textbf{89.57} \\
8 & 89.77 & 77.10 & 85.53 & 84.42 & \underline{93.33} & 92.96 & 90.69 & 78.87 & 24.26 & 5.91 & \textbf{98.11} \\
9 & 84.40 & \textbf{98.48} & 82.64 & 93.03 & 97.14 & 81.02 & 91.02 & 65.62 & 9.01 & 1.66 & \underline{97.20} \\
10 & 82.47 & \textbf{91.91} & 61.00 & 86.16 & \underline{87.58} & 46.08 & 70.42 & 39.27 & 14.11 & 0.46 & 86.42 \\
11 & 90.77 & \textbf{99.32} & 81.41 & 91.18 & \underline{94.78} & 88.12 & 80.75 & 70.29 & 18.31 & 1.72 & 93.19 \\
12 & 67.64 & \textbf{91.80} & 52.87 & 63.56 & 85.45 & 60.61 & 73.66 & 55.59 & 28.50 & 3.21 & \underline{87.75} \\
13 & \underline{85.99} & 50.80 & 45.48 & 35.34 & 29.04 & 12.91 & 30.22 & 32.89 & 35.78 & 6.01 & \textbf{91.60} \\
14 & 93.52 & 54.56 & 15.81 & 22.79 & \underline{94.12} & 38.21 & 68.54 & 15.10 & 15.08 & 5.54 & \textbf{98.49} \\
15 & 36.62 & 30.93 & 37.71 & \textbf{43.70} & \underline{42.78} & 40.65 & 32.64 & 29.70 & 16.68 & 0.63 & 28.96 \\
16 & 95.62 & 65.11 & 90.13 & 94.97 & 97.47 & 95.32 & 84.79 & \underline{98.27} & 17.27 & 9.58 & \textbf{99.01} \\
17 & 60.54 & \underline{95.74} & 56.24 & 91.71 & 78.72 & 59.11 & 82.54 & 13.29 & 7.33 & -0.10 & \textbf{96.44} \\
18 & 63.92 & 17.39 & 58.47 & \underline{70.21} & 54.61 & 47.93 & 64.15 & 10.01 & 3.79 & 0.65 & \textbf{79.70} \\
19 & 62.12 & 79.06 & 49.36 & 65.77 & \underline{81.07} & 51.19 & 63.41 & 26.84 & 2.73 & 1.06 & \textbf{84.13} \\
\midrule
Mean & 75.04 & 75.88 & 60.45 & 71.29 & \underline{79.94} & 61.26 & 66.78 & 49.17 & 15.66 & 2.48 & \textbf{86.33} \\
Mean-Rank & 4.80 & 3.50 & 6.85 & 4.50 & \underline{3.20} & 6.35 & 6.05 & 7.80 & 9.75 & 11.00 & \textbf{2.20} \\
Top-3 & 4 & 14 & 0 & 6 & \underline{15} & 2 & 1 & 1 & 0 & 0 & \textbf{17} \\
Top-1 & 0 & \underline{8} & 0 & 1 & 1 & 0 & 0 & 0 & 0 & 0 & \textbf{10} \\
\bottomrule
\end{tabular}
\end{table}

\FloatBarrier

\subsection{Are \our{}'s assignments well calibrated?}
\label{appendix:extended_results_brier}

Tables~\ref{tab:extended_real_brier}, \ref{tab:extended_gauss_brier} and \ref{tab:extended_transf_brier} contain detailed extension of Table~\ref{tab:comp_brier} from the main part of the paper. Similar to the Tables in Appendix~\ref{appendix:extended_results_ari}, these also include additional rows: Mean, Mean-Rank, Top-3, and Top-1, which aggregate the results presented in each table.

\our{} consistently attains at least the second position across all statistics in every table presented in this section. For the Synthetic Transformed datasets, it is undeniably the best. However, for the OpenML datasets, it is outperformed by IDEC, and for the Synthetic Gaussian datasets, GMM takes the lead in the Top-1 metric, while for the other statistics, k-means performs better.

\begin{table}[!ht]
    \footnotesize
    \centering
    \caption{Soft clustering performance of ZEUS versus competing methods on real-world datasets, measured by Brier score (lower is better).}
    \label{tab:extended_real_brier}
    \begin{tabular}{lcccccccc}
        \toprule
        ID & KM & GMM & AE-KM & AE-GMM & DEC & IDEC & G-CEALS & \our \\
        \midrule
        14 & 0.7660 & 0.8579 & 0.8240 & 0.7144 & \textbf{0.5338} & \underline{0.6409} & 0.7993 & 0.7006 \\
        15 & \underline{0.0858} & 0.1548 & 0.2100 & 0.1791 & 0.1489 & \textbf{0.0644} & 0.2003 & 0.0973 \\
        16 & 0.5012 & 0.4982 & 0.4224 & 0.5372 & \underline{}{0.3726} & 0.5020 & 0.6099 & \textbf{0.2761} \\
        18 & \underline{0.6060} & 0.8529 & 0.7312 & 0.6961 & 0.6099 & \textbf{0.5354} & 0.6855 & 0.7094 \\
        22 & 0.8316 & 0.7533 & 1.0686 & 0.7684 & \textbf{0.5071} & 0.\underline{6496} & 0.7201 & 0.7173 \\
        35 & \textbf{0.0710} & 0.5016 & 0.4678 & 0.5711 & 0.5881 & 0.3987 & 0.3836 & \underline{0.2552} \\
        51 & 0.3741 & 0.3741 & 0.3687 & 0.3844 & \textbf{0.3297} & 0.3345 & 0.3997 & \underline{0.3333} \\
        53 & 0.4074 & 0.7052 & 0.3556 & 0.6285 & \underline{0.3361} & \textbf{0.2979} & 0.5234 & 0.4000 \\
        56 & 0.2391 & 0.2344 & 0.2520 & 0.2703 & 0.2147 & \underline{0.2088} & 0.2517 & \textbf{0.1829} \\
        61 & 0.2267 & \textbf{0.0570} & 0.3600 & 0.3796 & 0.3881 & 0.3445 & 0.4167 & \underline{0.0979} \\
        187 & 0.0899 & \textbf{0.0622} & 0.1349 & 0.1146 & 0.3380 & 0.0785 & 0.2518 & \underline{0.0762} \\
        377 & 0.8633 & 0.8320 & 0.6200 & 0.6275 & \underline{0.5316} & \textbf{0.5047} & 0.5739 & 0.8508 \\
        458 & 0.0285 & \underline{0.0214} & 0.5222 & 0.3407 & 0.3203 & 0.3656 & 0.4104 & \textbf{0.0048} \\
        481 & 0.8038 & 0.8134 & 0.4536 & \textbf{0.2136} & 0.3111 & \underline{0.2146} & 0.4422 & 0.6986 \\
        694 & 1.0155 & 0.8831 & 0.9858 & 0.9342 & 0.7616 & \underline{0.7525} & \textbf{0.7038} & 0.7621 \\
        721 & \textbf{0.3400} & \textbf{0.3400} & 0.6620 & 0.6031 & 0.4491 & \underline{0.4359} & 0.7937 & \textbf{0.3400} \\
        733 & 0.1722 & 0.2859 & 0.3082 & 0.4937 & 0.2115 & \textbf{0.1129} & 0.3195 & \underline{0.1244} \\
        745 & 0.2516 & 0.2516 & \textbf{0.1359} & 0.2766 & 0.3016 & 0.1793 & 0.2388 & \underline{0.1384} \\
        756 & 0.2767 & 0.3615 & \underline{0.2113} & 0.3262 & 0.3178 & \textbf{0.1329} & 0.4134 & 0.2767 \\
        796 & \textbf{0.0574} & 0.7943 & \underline{0.1340} & 0.3088 & 0.2046 & 0.3002 & 0.5269 & 0.6253 \\
        820 & 0.2979 & 0.3720 & 0.3557 & 0.2876 & \textbf{0.2334} & \underline{0.2594} & 0.3981 & 0.4596 \\
        840 & 0.8780 & 0.2790 & 0.2868 & \underline{0.2625} & 0.2746 & \textbf{0.2445} & 0.4642 & 0.4665 \\
        854 & 0.6329 & 0.8203 & 0.3468 & 0.4481 & 0.4469 & \underline{0.3372} & 0.3678 & \textbf{0.1266} \\
        1462 & 0.8484 & 0.8978 & 0.7688 & 0.8310 & \underline{0.6674} & 0.7714 & 0.6747 & \textbf{0.0394} \\
        1495 & \underline{0.0160} & \textbf{0.0080} & 0.1696 & 0.0397 & 0.2667 & 0.0713 & 0.1583 & 0.7200 \\
        1499 & 0.2190 & 0.2933 & 0.1867 & 0.1961 & 0.3939 & \underline{0.1376} & 0.4063 & \textbf{0.1239} \\
        1510 & 0.1441 & \textbf{0.1161} & 0.2257 & 0.1954 & 0.2195 & 0.1432 & 0.2060 & \underline{0.1371} \\
        1523 & 1.0516 & 0.7284 & 0.9019 & 0.9708 & \textbf{0.6099} & 0.7782 & \underline{0.6316} & 1.0002 \\
        4153 & 0.7778 & 0.7178 & 0.6311 & 0.8111 & \underline{0.5757} & 0.6377 & 0.8360 & \textbf{0.4651} \\
        40496 & \textbf{0.5064} & 0.9563 & 0.8312 & 0.8432 & 0.6733 & \underline{0.6143} & 0.7376 & 1.0124 \\
        40682 & 0.2233 & \underline{0.0747} & 0.2456 & 0.2251 & 0.2694 & \textbf{0.0685} & 0.3210 & 0.2884 \\
        40705 & 0.3879 & 0.3837 & 0.7095 & 0.4415 & \textbf{0.2541} & 0.3347 & 0.4304 & \underline{0.3254} \\
        42261 & 0.2267 & \textbf{0.0570} & 0.3413 & 0.3760 & 0.3889 & 0.2854 & 0.3800 & \underline{0.1081} \\
        42585 & 0.6279 & 0.9790 & 0.5558 & 0.6111 & \underline{0.3494} & 0.7425 & 0.3795 & \textbf{0.0383} \\
        \midrule
        Mean & 0.4366 & 0.4799 & 0.4643 & 0.4679 & 0.3941 & \textbf{0.3671} & 0.4722 & \underline{0.3817} \\
        Mean-Rank & 4.54 & 4.94 & 5.12 & 5.50 & 3.88 & \textbf{2.85} & 5.65 & \underline{3.51} \\
        Top-3 & 12 & 9 & 8 & 4 & 19 & \textbf{22} & 6 & \underline{21} \\
        Top-1 & 3 & 5 & 1 & 1 & \underline{6} & \textbf{8} & 1 & \textbf{8} \\
        \bottomrule
    \end{tabular}
\end{table}

\begin{table}[!ht]
    \footnotesize
    \centering
    \caption{Soft clustering performance (Brier score) of ZEUS compared to baseline methods on Synthetic Gaussian datasets (lower score reflect better quality).}
    \label{tab:extended_gauss_brier}
    \begin{tabular}{lcccccccc}
        \toprule
        ID & KM & GMM & AE-KM & AE-GMM & DEC & IDEC & G-CEALS & \our \\
        \midrule
        0 & \underline{0.1448} & 0.2200 & 0.1741 & 0.3558 & 0.4251 & 0.3264 & 0.3921 & \textbf{0.1242} \\
        1 & \textbf{0.1218} & \underline{0.1886} & 0.2337 & 0.4718 & 0.4379 & 0.4109 & 0.4377 & 0.4502 \\
        2 & 0.1985 & 0.9985 & 0.2995 & \textbf{0.1922} & 0.4303 & 0.2321 & 0.5400 & \underline{0.1952} \\
        3 & \underline{0.0528} & 0.9078 & 0.0927 & 0.3012 & 0.6196 & 0.1628 & 0.4194 & \textbf{0.0413} \\
        4 & 0.1005 & \textbf{0.0575} & 0.1793 & 0.3351 & 0.4955 & 0.2312 & 0.3886 & \underline{0.0792} \\
        5 & \underline{0.1020} & 1.0950 & 0.1333 & 0.4892 & 0.3355 & 0.2059 & 0.3995 & \textbf{0.0881} \\
        6 & 0.0350 & \textbf{0.0156} & 0.1200 & 0.0489 & 0.2971 & 0.0770 & 0.3561 & \underline{0.0209} \\
        7 & \underline{0.0632} & 0.9781 & 0.1599 & 0.2610 & 0.4693 & 0.1207 & 0.6146 & \textbf{0.0543} \\
        8 & \underline{0.1168} &\textbf{0.0844} & 0.2079 & 0.2786 & 0.4626 & 0.2970 & 0.5975 & 0.1631 \\
        9 & 0.0677 & \textbf{0.0092} & 0.3552 & 0.1925 & 0.2167 & 0.2090 & 0.4586 & \underline{0.0573} \\
        10 & 0.1091 & \textbf{0.0292} & 0.1469 & \underline{0.0460} & 0.3300 & 0.1229 & 0.3127 & 0.0655 \\
        11 & 0.0558 & \textbf{0.0117} & 0.0791 & 0.1370 & 0.4110 & 0.1122 & 0.4437 & \underline{0.0234} \\
        12 & \textbf{0.1361} & 0.2296 & 0.4022 & 0.3492 & 0.5917 & 0.3230 & 0.3583 & \underline{0.1545} \\
        13 & \underline{0.0216} & \textbf{0.0115} & 0.0356 & 0.0629 & 0.3223 & 0.0980 & 0.1947 & 0.0270 \\
        14 & \underline{0.1048} & 0.8000 & 0.3009 & 0.3776 & 0.3359 & 0.2380 & 0.3818 & \textbf{0.0626} \\
        15 & 0.0734 & \textbf{0.0278} & 0.2574 & 0.0746 & 0.3458 & 0.2483 & 0.3120 & \underline{0.0352} \\
        16 & \underline{0.0537} & \textbf{0.0448} & 0.0958 & 0.1747 & 0.1640 & 0.2528 & 0.1454 & 0.1770 \\
        17 & \underline{0.1309} & 0.3513 & 0.4178 & 0.2891 & 0.4313 & 0.2756 & 0.5425 & \textbf{0.1308} \\
        18 & \underline{0.1035} & \textbf{0.0452} & 0.1458 & 0.2319 & 0.5193 & 0.2117 & 0.2961 & 0.1939 \\
        19 & \underline{0.1475} & \textbf{0.1132} & 0.3094 & 0.2989 & 0.2444 & 0.4612 & 0.3015 & 0.3942 \\
        \midrule
        Mean & \textbf{0.0970} & 0.3110 & 0.2073 & 0.2484 & 0.3943 & 0.2308 & 0.3946 & \underline{0.1269} \\
        Mean-Rank & \textbf{2.30} & 3.25 & 4.55 & 4.95 & 6.65 & 4.85 & 6.75 & \underline{2.70} \\
        Top-3 & \textbf{19} & 13 & 6 & 2 & 1 & 3 & 0 & \underline{16} \\
        Top-1 & 2 & \textbf{11} & 0 & 1 & 0 & 0 & 0 & \underline{6} \\
        \bottomrule
    \end{tabular}
\end{table}

\begin{table}[!ht]
    \footnotesize
    \centering
    \caption{Evaluation of soft clustering quality on Synthetic Transformed datasets using Brier score (lower score indicate better performance).}
    \label{tab:extended_transf_brier}
    \begin{tabular}{lcccccccc}
        \toprule
        ID & KM & GMM & AE-KM & AE-GMM & DEC & IDEC & G-CEALS & \our \\
        \midrule
        0 & 0.5116 & \underline{0.2135} & 0.6639 & 0.4457 & 0.5911 & 0.5063 & 0.7116 & \textbf{0.1282} \\
        1 & \underline{0.0666} & \textbf{0.0484} & 0.2328 & 0.1514 & 0.4755 & 0.5045 & 0.2806 & 0.2718 \\
        2 & 0.0883 & \textbf{0.0330} & 0.4996 & 0.2822 & 0.4424 & 0.3441 & 0.5990 & \underline{0.0394} \\
        3 & 0.3254 & 0.6286 & 0.5872 & \textbf{0.2118} & 0.3241 & 0.4862 & 0.7537 & \underline{0.2678} \\
        4 & 0.6757 & \textbf{0.1547} & 0.5580 & 0.6123 & 0.3920 & 0.5158 & 0.3804 & \underline{0.3190} \\
        5 & \underline{0.0887} & \textbf{0.0359} & 0.3201 & 0.1517 & 0.6415 & 0.4114 & 0.3869 & 0.2802 \\
        6 & 0.5771 & \underline{0.1788} & 0.4789 & 0.4049 & 0.5718 & 0.3069 & 0.5448 & \textbf{0.1763} \\
        7 & 0.3634 & \underline{0.2875} & 0.4281 & 0.3326 & 0.5478 & 0.3379 & 0.6068 & \textbf{0.2461} \\
        8 & 0.1048 & 0.2738 & 0.1271 & 0.0817 & 0.4517 & \underline{0.0784} & 0.3278 & \textbf{0.0167} \\
        9 & 0.3014 & \textbf{0.0120} & 0.3120 & 0.0940 & 0.4474 & 0.3329 & 0.2284 & \underline{0.0481} \\
        10 & 0.1785 & \textbf{0.0763} & 0.3674 & \underline{0.1249} & 0.3845 & 0.5057 & 0.5420 & 0.1387 \\
        11 & \underline{0.1053} & \textbf{0.0130} & 0.1684 & 0.1128 & 0.6071 & 0.2580 & 0.3306 & 0.1768 \\
        12 & 0.3682 & \textbf{0.0657} & 0.6462 & 0.4740 & 0.3838 & 0.4585 & 0.6318 & \underline{0.1936} \\
        13 & \underline{0.0604} & 0.2549 & 0.4206 & 0.4292 & 0.3711 & 0.4073 & 0.3436 & \textbf{0.0355} \\
        14 & \underline{0.0445} & 0.7982 & 1.1164 & 0.6259 & 0.3672 & 0.4949 & 0.6715 & \textbf{0.0097} \\
        15 & 0.7142 & 0.7088 & 0.6256 & 0.7924 & 0.5801 & \underline{0.5741} & \textbf{0.4807} & 0.8854 \\
        16 & \underline{0.0524} & 0.4745 & 0.1194 & \textbf{0.0303} & 0.5832 & 0.1743 & 0.1014 & 0.0809 \\
        17 & 0.2586 & \underline{0.0230} & 0.2616 & 0.0394 & 0.3480 & 0.2156 & 0.6361 & \textbf{0.0177} \\
        18 & \underline{0.2081} & 0.6698 & 0.3781 & 0.2409 & 0.3774 & 0.3861 & 0.7879 & \textbf{0.1056} \\
        19 & 0.5120 & \underline{0.1821} & 0.7577 & 0.6415 & 0.3892 & 0.4845 & 0.5570 & \textbf{0.1542} \\
        \midrule
        Mean & 0.2803 & \underline{0.2566} & 0.4535 & 0.3140 & 0.4638 & 0.3892 & 0.4951 & \textbf{0.1796} \\
        Mean-Rank & 3.95 & \underline{3.00} & 5.85 & 3.95 & 5.75 & 5.00 & 6.15 & \textbf{2.35} \\
        Top-3 & 9 & \underline{14} & 0 & 12 & 4 & 3 & 2 & \textbf{16} \\
        Top-1 & 0 & \underline{8} & 0 & 2 & 0 & 0 & 1 & \textbf{9} \\
        \bottomrule
    \end{tabular}
\end{table}

\FloatBarrier

\subsection{ How helpful is regularization for \our{}?}
\label{appendix:extended_results_ablations}

Tables~\ref{tab:extended_real_ablations}, \ref{tab:extended_gauss_ablations} and \ref{tab:extended_transf_ablations} provide extended versions of Table~\ref{tab:comp_ablations} , which analyzes the impact of different combinations of regularization functions. Their structure is analogous to the other tables presented in Appendix~\ref{sec:exp_app}.

The conclusions that can be drawn from these extended tables, along with their statistical summaries, are consistent with our original claims. The model with both $\Loss_{sep}$ and $\Loss_{cp}$ regularizers performs best on real-world data collections and consistently ranks at least second in the considered statistics for synthetic data. On these generated datasets, it is frequently outperformed by the variant that includes the  $\Loss_{cp}$ component alone. By contrast, the remaining two approaches clearly grapple with proper clustering of the synthetic datasets.

\begin{table}[!ht]
    \footnotesize
    \centering
    \caption{Impact of regularization components on OpenML datasets, evaluated using the ARI score (higher is better).}
    \label{tab:extended_real_ablations}
    \begin{tabular}{lcccc}
        \toprule
        ID & $\Loss_{prob}$ & $\Loss_{prob} + \Loss_{sep}$ & $\Loss_{prob} + \Loss_{cp}$ & $\Loss_{prob} + \Loss_{sep} + \Loss_{cp}$ \\
        \midrule
        14 & \underline{48.40} & 43.49 & 46.44 & \textbf{50.56} \\
        15 & \underline{86.64} & \textbf{87.17} & 82.94 & 81.28 \\
        16 & 69.87 & 73.33 & \textbf{81.87} & \underline{74.03} \\
        18 & 43.06 & \underline{50.80} & 49.62 & \textbf{51.63} \\
        22 & \underline{63.27} & 56.86 & \textbf{65.58} & 56.05 \\
        35 & 52.33 & 74.58 & \underline{76.90} & \textbf{85.12} \\
        51 & 32.31 & 31.54 & \underline{41.84} & \textbf{43.66} \\
        53 & 5.17 & \underline{27.38} & 25.85 & \textbf{35.76} \\
        56 & 58.48 & \underline{61.34} & 20.84 & \textbf{66.41} \\
        61 & 48.93 & \textbf{88.57} & 65.37 & \underline{85.15} \\
        187 & \underline{88.38} & \textbf{91.50} & 88.22 & 88.19 \\
        377 & 46.31 & 45.16 & \underline{54.56} & \textbf{55.30} \\
        458 & 98.16 & \underline{98.49} & \underline{98.49} & \textbf{99.19} \\
        481 & 2.77 & \textbf{25.87} & 1.53 & \underline{8.68} \\
        694 & \underline{42.76} & \textbf{48.86} & 30.35 & 33.38 \\
        721 & \textbf{43.28} & \underline{5.39} & \textbf{43.28} & \textbf{43.28} \\
        733 & 34.35 & 24.94 & \textbf{76.83} & \underline{74.97} \\
        745 & \underline{59.56} & 22.80 & 4.46 & \textbf{73.85} \\
        756 & \textbf{76.15} & \underline{63.49} & 38.42 & 50.31 \\
        796 & 6.56 & 3.96 & \textbf{14.52} & \underline{13.99} \\
        820 & \textbf{39.95} & \textbf{39.95} & \underline{34.72} & 28.87 \\
        840 & 0.95 & 4.13 & \textbf{31.14} & \underline{27.66} \\
        854 & 12.37 & 12.37 & \underline{24.32} & \textbf{76.14} \\
        1462 & 92.00 & \textbf{94.25} & 62.85 & \underline{92.28} \\
        1495 & -0.29 & \textbf{98.40} & \underline{7.35} & \underline{7.35} \\
        1499 & 60.63 & 44.86 & \underline{71.34} & \textbf{82.40} \\
        1510 & 70.09 & \underline{73.67} & 72.51 & \textbf{74.26} \\
        1523 & \underline{25.68} & \textbf{28.13} & 17.38 & 19.60 \\
        4153 & 38.28 & \underline{51.26} & 50.24 & \textbf{62.29} \\
        40496 & \textbf{34.15} & \underline{33.63} & 30.68 & 32.25 \\
        40682 & -0.39 & 23.38 & \underline{42.79} & \textbf{53.15} \\
        40705 & 40.00 & \underline{40.51} & 38.18 & \textbf{45.45} \\
        42261 & 48.93 & \textbf{88.57} & 65.37 & \underline{85.15} \\
        42585 & 53.97 & \underline{95.82} & \textbf{97.39} & 95.05 \\
        \midrule
        Mean & 44.80 & \underline{51.60} & 48.65 & \textbf{57.43} \\
        Mean-rank & 3.00 & \underline{2.37} & 2.68 & \textbf{1.96} \\
        Top-3 & 20 & \underline{26} & 25 & \textbf{30} \\
        Top-1 & 2 & \underline{9} & 6 & \textbf{15} \\
        \bottomrule
    \end{tabular}
\end{table}

\begin{table}[!ht]
    \footnotesize
    \centering
    \caption{Effect of regularization components on Synthetic Gaussian datasets, assessed by the ARI metric (higher score indicate better quality).}
    \label{tab:extended_gauss_ablations}
    \begin{tabular}{lcccc}
        \toprule
        ID & $\Loss_{prob}$ & $\Loss_{prob} + \Loss_{sep}$ & $\Loss_{prob} + \Loss_{cp}$ & $\Loss_{prob} + \Loss_{sep} + \Loss_{cp}$ \\
        \midrule
        0 & 72.89 & 68.81 & \underline{85.85} & \textbf{87.65} \\
        1 & \textbf{89.50} & 71.68 & \underline{88.72} & 72.04 \\
        2 & 80.18 & 63.23 & \textbf{92.59} & \underline{82.96} \\
        3 & 77.05 & 79.22 & \underline{91.23} & \textbf{95.76} \\
        4 & 72.32 & 84.96 & \underline{92.82} & \textbf{92.85} \\
        5 & \textbf{91.88} & 88.98 & 82.25 & \underline{90.72} \\
        6 & 81.85 & 81.18 & \underline{87.49} & \textbf{97.89} \\
        7 & 85.33 & 79.93 & \textbf{93.75} & \underline{93.60} \\
        8 & 72.54 & 76.12 & \textbf{87.92} & \underline{84.00} \\
        9 & \underline{95.47} & \textbf{95.97} & 93.59 & 94.55 \\
        10 & \underline{88.23} & \textbf{89.61} & 85.07 & 87.32 \\
        11 & \underline{97.26} & 94.91 & 96.55 & \textbf{97.45} \\
        12 & 77.58 & 78.07 & \underline{85.81} & \textbf{86.04} \\
        13 & 74.41 & 76.08 & \underline{96.07} & \textbf{96.17} \\
        14 & 84.00 & 82.11 & \textbf{95.39} & \underline{94.60} \\
        15 & 77.65 & 78.94 & \textbf{97.10} & \underline{96.68} \\
        16 & \underline{93.82} & 93.46 & \textbf{95.82} & 87.19 \\
        17 & \textbf{88.83} & 86.38 & \underline{88.05} & 86.29 \\
        18 & 89.90 & \textbf{90.21} & \underline{90.02} & 83.01 \\
        19 & 76.62 & \underline{77.69} & \textbf{85.63} & 73.85 \\
        \midrule
        Mean & 83.37 & 81.88 & \textbf{90.59} & \underline{89.03} \\
        Mean-rank & 2.80 & 3.00 & \textbf{2.00} & \underline{2.20} \\
        Top-3 & 14 & 13 & \textbf{17} & \underline{16} \\
        Top-1 & \underline{3} & \underline{3} & \textbf{7} & \textbf{7} \\
        \bottomrule
    \end{tabular}
\end{table}

\begin{table}[!ht]
    \footnotesize
    \centering
    \caption{Regularisation impact on Synthetic Transformed datasets, measured by ARI (higher is better).}
    \label{tab:extended_transf_ablations}
    \begin{tabular}{lcccc}
        \toprule
        ID & $\Loss_{prob}$ & $\Loss_{prob} + \Loss_{sep}$ & $\Loss_{prob} + \Loss_{cp}$ & $\Loss_{prob} + \Loss_{sep} + \Loss_{cp}$ \\
        \midrule
        0 & 67.29 & 71.62 & \underline{85.76} & \textbf{87.42} \\
        1 & \underline{91.98} & 91.46 & \textbf{96.88} & 89.40 \\
        2 & 93.17 & \underline{93.84} & 92.36 & \textbf{95.88} \\
        3 & 68.26 & 71.68 & \underline{80.53} & \textbf{80.98} \\
        4 & 80.93 & \underline{81.64} & \textbf{84.22} & 73.66 \\
        5 & 81.38 & 80.06 & \textbf{95.99} & \underline{81.63} \\
        6 & \underline{84.29} & 82.06 & 80.67 & \textbf{87.02} \\
        7 & 83.35 & 73.07 & \textbf{96.83} & \underline{89.57} \\
        8 & 97.42 & \textbf{98.61} & 97.16 & \underline{98.11} \\
        9 & \textbf{97.63} & 96.43 & 96.37 & \underline{97.20} \\
        10 & \textbf{91.32} & 77.37 & \underline{90.58} & 86.42 \\
        11 & 91.84 & 89.07 & \textbf{99.17} & \underline{93.19} \\
        12 & 74.93 & \textbf{92.20} & \underline{91.14} & 87.75 \\
        13 & 0.32 & 0.73 & \underline{90.67} & \textbf{91.60} \\
        14 & 98.00 & \textbf{98.83} & 98.16 & \underline{98.49} \\
        15 & \underline{43.28} & \textbf{48.54} & 30.50 & 28.96 \\
        16 & 93.34 & 90.38 & \underline{98.11} & \textbf{99.01} \\
        17 & \underline{97.62} & \textbf{98.21} & 95.86 & 96.44 \\
        18 & \underline{82.64} & 82.22 & \textbf{86.94} & 79.70 \\
        19 & 77.93 & 67.71 & \underline{83.78} & \textbf{84.13} \\
        \midrule
        Mean & 79.85 & 79.29 & \textbf{88.58} & \underline{86.33} \\
        Mean-rank & 2.80 & 2.70 & \underline{2.30} & \textbf{2.20} \\
        Top-3 & 15 & 14 & \underline{15} & \textbf{16} \\
        Top-1 & 2 & 5 & \underline{6} & \textbf{7} \\
    \bottomrule
    \end{tabular}
\end{table}

\FloatBarrier

\subsection{What data-generating prior is optimal?}

Tables~\ref{tab:extended_prior_real}, \ref{tab:extended_prior_gauss}, and \ref{tab:extended_prior_transf} expand upon the findings presented in Table~\ref{tab:prior_ablation} by examining the influence of various data-generating priors across all datasets within each group. Their format is consistent with other tables in Appendix~\ref{sec:exp_app}.

\begin{table}[ht!]
\footnotesize
\caption{Evaluation of clustering quality with the ARI metric on real-world datasets for different combinations of data-generating probabilistic models (higher scores reflect better performance).}
\centering
\label{tab:extended_prior_real}
\begin{tabular}{lcccc}
\toprule
ID & Gauss. + Cat. & NN-transf. + Cat. & Gauss. + NN-transf. & Gauss. + NN-transf. + Cat.\\
\midrule
14 & 36.81 & \underline{49.84} & 49.84 & \textbf{50.56} \\
15 & \underline{85.52} & \textbf{85.57} & 83.46 & 81.28 \\
16 & 57.15 & 73.05 & \textbf{76.01} & \underline{74.03} \\
18 & 42.83 & 46.48 & \textbf{55.29} & \underline{51.63} \\
22 & 46.45 & \textbf{62.23} & \underline{60.63} & 56.02 \\
35 & \textbf{87.48} & 76.48 & 84.25 & \underline{85.12} \\
51 & -0.52 & 38.26 & \underline{39.76} & \textbf{43.66} \\
53 & \textbf{37.57} & 29.79 & 27.38 & \underline{35.76} \\
56 & 48.40 & 56.35 & \underline{59.92} & \textbf{66.41} \\
61 & 68.44 & 62.26 & \underline{85.08} & \textbf{85.15} \\
187 & \underline{89.77} & \textbf{94.87} & 88.22 & 88.19 \\
377 & 35.12 & \underline{58.34} & \textbf{58.95} & 55.30 \\
458 & 47.43 & \underline{98.79} & 98.47 & \textbf{99.19} \\
481 & 6.46 & \underline{6.94} & 3.58 & \textbf{8.68} \\
694 & \underline{27.75} & 19.90 & 27.69 & \textbf{33.38} \\
721 & \textbf{43.28} & \underline{43.28} & 38.13 & 43.28 \\
733 & 73.48 & -7.69 & \textbf{75.77} & \underline{74.97} \\
745 & 4.46 & \underline{67.51} & 28.11 & \textbf{73.85} \\
756 & 4.46 & \textbf{59.53} & 4.46 & \underline{50.31} \\
796 & \underline{14.80} & 5.94 & \textbf{15.29} & 13.99 \\
820 & 19.26 & \textbf{36.77} & \underline{33.71} & 28.87 \\
840 & -0.83 & 0.69 & \underline{2.57} & \textbf{27.66} \\
854 & 24.32 & \underline{73.92} & 12.56 & \textbf{76.14} \\
1462 & 1.92 & \textbf{92.56} & \underline{92.56} & 92.28 \\
1495 & -0.29 & 0.03 & \textbf{55.17} & \underline{7.35} \\
1499 & 58.62 & \underline{74.43} & 71.07 & \textbf{82.40} \\
1510 & 72.50 & \underline{72.95} & 70.08 & \textbf{74.26} \\
1523 & \underline{20.38} & 18.69 & \textbf{22.01} & 19.60 \\
4153 & 41.49 & 46.33 & \underline{56.13} & \textbf{62.29} \\
40496 & \textbf{33.20} & 31.69 & 30.61 & \underline{32.25} \\
40682 & \underline{64.49} & 58.59 & \textbf{77.21} & 53.15 \\
40705 & 29.52 & \underline{39.74} & 37.15 & \textbf{45.45} \\
42261 & 66.34 & 62.26 & \underline{81.76} & \textbf{85.15} \\
42585 & 92.06 & \underline{94.29} & 64.95 & \textbf{95.05} \\
\midrule
Mean & 40.59 & 50.90 & \underline{52.00} & \textbf{57.43} \\
Mean-Rank & 3.10 & 2.56 & \underline{2.46} & \textbf{1.88} \\
Top-3 & 17 & \underline{27} & 26 & \textbf{31} \\
Top-1 & 3 & 5 & \underline{8} & \textbf{16} \\
\bottomrule
\end{tabular}
\end{table}

\begin{table}[ht!]
\footnotesize
\caption{Effect of prior selection on \our{}'s ARI performance on Synthetic Gaussian datasets (higher scores indicate better performance).}
\centering
\label{tab:extended_prior_gauss}
\begin{tabular}{lcccc}
\toprule
ID & Gauss. + Cat. & NN-transf. + Cat. & Gauss. + NN-transf. & Gauss. + NN-transf. + Cat.\\
\midrule
0 & \textbf{88.78} & 87.09 & 82.89 & \underline{87.65} \\
1 & \textbf{87.74} & 79.51 & \underline{80.38} & 72.04 \\
2 & \textbf{91.30} & \underline{87.30} & 20.50 & 82.96 \\
3 & \underline{95.08} & 93.27 & 44.56 & \textbf{95.76} \\
4 & \underline{94.40} & \textbf{94.46} & 92.48 & 92.85 \\
5 & \textbf{91.43} & 90.08 & 30.45 & \underline{90.72} \\
6 & \textbf{98.05} & 97.22 & \underline{98.02} & 97.89 \\
7 & \textbf{94.56} & \underline{94.43} & 26.90 & 93.60 \\
8 & 87.65 & \underline{87.95} & \textbf{88.05} & 84.00 \\
9 & \textbf{95.92} & 92.70 & \underline{94.79} & 94.55 \\
10 & \textbf{90.99} & 86.78 & \underline{90.07} & 87.32 \\
11 & \underline{96.86} & 96.52 & 96.74 & \textbf{97.45} \\
12 & \textbf{88.21} & \underline{86.25} & 85.55 & 86.04 \\
13 & \underline{96.07} & 93.39 & 93.81 & \textbf{96.17} \\
14 & \textbf{95.42} & 92.12 & 32.24 & \underline{94.60} \\
15 & \textbf{97.71} & 95.84 & \underline{97.14} & 96.68 \\
16 & \underline{95.24} & 94.49 & \textbf{95.50} & 87.19 \\
17 & \textbf{89.13} & 84.54 & \underline{86.66} & 86.29 \\
18 & \textbf{90.14} & 87.66 & \underline{89.37} & 83.01 \\
19 & \textbf{87.43} & 76.30 & \underline{79.00} & 73.85 \\
\midrule
Mean & \textbf{92.61} & \underline{89.90} & 75.25 & 89.03 \\
Mean-Rank & \textbf{1.35} & 3.05 & \underline{2.80} & \underline{2.80} \\
Top-3 & \textbf{20} & 13 & 12 & \underline{15} \\
Top-1 & \textbf{14} & 1 & 2 & \underline{3} \\
\bottomrule
\end{tabular}
\end{table}

\begin{table}[ht!]
\footnotesize
\caption{Influence of the chosen data-generating prior on \our{} performance over Synthetic Transformed datasets, reported in terms of ARI (larger values denote better clustering quality).}
\centering
\label{tab:extended_prior_transf}
\begin{tabular}{lcccc}
\toprule
ID & Gauss. + Cat. & NN-transf. + Cat. & Gauss. + NN-transf. & Gauss. + NN-transf. + Cat.\\
\midrule
0 & 61.54 & 87.33 & \textbf{88.33} & \underline{87.42} \\
1 & 84.60 & \textbf{95.73} & \underline{94.60} & 89.40 \\
2 & 83.73 & 93.89 & \textbf{97.59} & \underline{95.88} \\
3 & 45.52 & \underline{75.48} & 52.79 & \textbf{80.98} \\
4 & \underline{76.52} & 73.72 & \textbf{81.24} & 73.66 \\
5 & 72.95 & \textbf{83.07} & 79.33 & \underline{81.63} \\
6 & 80.87 & \underline{83.20} & 83.19 & \textbf{87.02} \\
7 & 88.97 & \textbf{95.91} & \underline{91.18} & 89.57 \\
8 & 97.55 & \underline{98.09} & 36.45 & \textbf{98.11} \\
9 & 94.25 & 96.86 & \textbf{98.29} & \underline{97.20} \\
10 & 40.64 & 81.56 & \textbf{88.46} & \underline{86.42} \\
11 & 93.42 & \underline{98.57} & \textbf{98.98} & 93.19 \\
12 & 83.00 & \underline{90.91} & \textbf{92.21} & 87.75 \\
13 & 80.76 & 90.87 & \textbf{94.12} & \underline{91.60} \\
14 & 98.19 & \textbf{98.49} & 13.51 & \underline{98.49} \\
15 & 28.32 & 28.47 & \textbf{29.06} & \underline{28.96} \\
16 & 98.40 & \textbf{99.05} & 18.65 & \underline{99.01} \\
17 & 87.75 & \textbf{98.21} & \underline{98.21} & 96.44 \\
18 & 5.50 & \textbf{83.07} & 3.79 & \underline{79.70} \\
19 & 64.25 & \textbf{88.36} & \underline{85.72} & 84.13 \\
\midrule
Mean & 73.34 & \textbf{87.04} & 71.29 & \underline{86.33} \\
Mean-Rank & 3.65 & \textbf{2.00} & \underline{2.08} & 2.27 \\
Top-3 & 6 & \textbf{20} & 16 & \underline{18} \\
Top-1 & 0 & \underline{6} & \textbf{9} & 3 \\
\bottomrule
\end{tabular}
\end{table}

\FloatBarrier

\section{Additional experiments}
\label{aappendix:additional_exp}
In this section, we present additional ablation studies and experiments related to \our{}, including analyses regarding its robustness, architectural choices, clustering strategies and performance in special-case scenarios.

\subsection{Robustness to noise and outliers}
In order to assess the robustness of our model under noise and outlier conditions, we consider two alternative evaluation setups:
\begin{itemize}
    \item \textbf{ZEUS + noise}, where we add Gaussian noise with a standard deviation of 0.05 to each data point before the ZEUS forward pass,
    \item \textbf{ZEUS + anomalies}, where we introduce a set of global anomalies corresponding to 5$\%$ of the dataset size. Each anomaly is generated by uniformly sampling numerical features from the range $(-1,1)$ and randomly selecting categorical values. The resulting dataset, containing both ground truth data points and anomalies, is then passed through our model. Final clustering quality is measured only on the ground truth points, for which the true class assignments are known.
\end{itemize}
Table~\ref{tab:disruptions} displays the outcomes of these experiments, alongside the performance of the vanilla ZEUS model.

\begin{table}[!ht]
    \centering
    \caption{Impact of Gaussian noise and anomalies on clustering performance.}
    \label{tab:disruptions}
    \begin{tabular}{llccc}
    \toprule
        Metric & Dataset group & ZEUS+noise & ZEUS+anomalies & ZEUS \\
        \midrule
        ARI	& Real & \underline{55.88} & 55.83 & \textbf{57.43} \\
        ARI	& Syn. Gauss. & 88.01 & \underline{88.21} & \textbf{89.03} \\
        ARI	& Syn. Transf. & 84.82 & \underline{85.17} & \textbf{86.33} \\
        \midrule
        Rank & Real & 2.24 & \underline{1.93} & \textbf{1.84} \\
        Rank & Syn. Gauss. & 2.45 & \underline{2.10} & \textbf{1.45} \\
        Rank & Syn. Transf. & 2.45 & \underline{2.30} & \textbf{1.25} \\
    \bottomrule
    \end{tabular}
\end{table}

\FloatBarrier
Overall, the average performance in the perturbed setups is not significantly lower than that of the vanilla ZEUS model, highlighting a degree of stability of our method in the presence of noise and outliers.

\subsection{Impact of token dimension}
The choice of token dimension also has a substantial influence on the performance and generalizability of \our{}. This is confirmed by an experiment in which we evaluated dimensions of 32, 128, 512, and 768 while keeping the standard setup unchanged. Table~\ref{tab:token_dim} reports the clustering performance of these variants, denoted as \emph{ZEUS-32}, \emph{ZEUS-128}, \emph{ZEUS-512}, and \emph{ZEUS-768}, respectively.

\begin{table}[!ht]
    \centering
    \caption{Assessment of different token dimensions in the \our{} architecture.}
    \begin{tabular}{llcccc}
        \toprule
        Metric & Dataset group & ZEUS-32 & ZEUS-128	& ZEUS-512 & ZEUS-768 \\
        \midrule
        ARI	& Real & 44.10 & 47.81 & \textbf{57.43} & \underline{50.79} \\
        ARI	& Syn. Gauss. & 69.50 & 83.04 & \underline{89.03} & \textbf{91.92} \\
        ARI	& Syn. Transf. & 67.57 & 81.62 & \underline{86.33} & \textbf{87.21} \\
        \midrule
        Rank & Real	& 2.90 & 2.96 & \textbf{1.90} & \underline{2.25} \\
        Rank & Syn. Gauss. & 4.00	& 3.00 & \underline{1.78} & \textbf{1.23} \\
        Rank & Syn. Transf.	& 3.90 & 2.78 & \underline{1.70} & \textbf{1.63} \\
        \bottomrule
    \end{tabular}
    \label{tab:token_dim}
\end{table}
\FloatBarrier

As one can infer, increasing the dimensionality leads to a consistent improvement in performance for synthetic data, whereas for OpenML datasets, there is a drop between dimensions 512 and 768, which may be caused by overfitting to the prior.

\subsection{Effect of the output clustering method}

Since \our{} is a method for learning representations suitable for clustering, in principle, any clustering algorithm could be used to separate the resulting clusters. However, our approach has the additional feature of being inherently designed to form spherical clusters; therefore, $K$-Means (\emph{ZEUS-KM}) should, in theory, already be sufficient. The primary advantage of $K$-means lies in its speed, which is a crucial factor for us during inference. Moreover, as demonstrated in the Table~\ref{tab:clust_meth_abl}, its clustering quality (measured by ARI) is comparable to the more computationally intensive Gaussian Mixture Model (\emph{ZEUS-GMM}). For reference, we also include results from a simplified variant of GMM with fixed identity covariance matrices (\emph{ZEUS-SGMM}), as employed in Section~\ref{sec:brier_exp}. All experiments are conducted using the same checkpoint and datasets as in Section~\ref{sec:quantitative_results}. 

\begin{table}[!ht]
    \centering
    \caption{Evaluation of clustering methods applied to the learned \our{} representations.}
    \begin{tabular}{llccc}
        \toprule
        Metric & Dataset group	& ZEUS-KM & ZEUS-SGMM & ZEUS-GMM \\
        \midrule
        ARI & Real & \textbf{57.43} & \underline{57.33} & 56.45 \\
        ARI & Syn. Gauss. & \underline{89.03} & 88.18 & \textbf{89.46} \\
        ARI & Syn. Transf. & \underline{86.33} & 85.57 & \textbf{87.22} \\
        \midrule
        Rank & Real & \textbf{1.94} & 2.04 & \underline{2.01} \\
        Rank & Syn. Gauss. & \underline{1.90} & 2.33 & \textbf{1.78} \\
        Rank & Syn. Transf. & \underline{2.13} & 2.35 & \textbf{1.53} \\
        \bottomrule
    \end{tabular}
    \label{tab:clust_meth_abl}
\end{table}
\FloatBarrier

Analysis of the average ARI scores indicates that ZEUS's output is well-suited to the evaluated approaches, as there are no significant differences in their results. Deep clustering techniques such as DEC or IDEC could also potentially be employed. Nevertheless, their reliance on intensive, dataset-specific training contradicts ZEUS's objective of delivering efficient, zero-shot representations.

\subsection{Integrating TabICL transformer into \our{}}
To mitigate one of the main limitations of our approach, namely the restricted number of input features and samples, we investigate whether it is feasible to replace the original TabPFN transformer architecture with a more recent design, such as the one used in TabICL~\cite{tabicl}. 
Specifically, we adopt the TabICL transformer with its default hyperparameters and integrate it into our unsupervised pipeline (\emph{\our{} + TabICL}) in exactly the same manner as described in Section~\ref{sec:setup}. The resulting clustering quality, alongside a comparison with the original \our{} model, are presented in Table~\ref{tab:icl_base}.

\begin{table}[!ht]
    \centering
    \caption{Performance of \our{} vs. its variant with the TabICL transformer backbone.}
    \label{tab:icl_base}
    \begin{tabular}{llcc}
        \toprule
        Metric & Dataset group & ZEUS & ZEUS + TabICL \\
        \midrule
         ARI	& Real & \textbf{57.43} & \underline{52.62} \\
         ARI	& Syn. Gauss. & \underline{89.03} & \textbf{89.81} \\
         ARI	& Syn. Transf. & \textbf{86.33} & \underline{85.97} \\
         \midrule
         Rank & Real	& \textbf{1.38} & \underline{1.62} \\
         Rank & Syn. Gauss. & \underline{1.60} & \textbf{1.40} \\
         Rank & Syn. Transf.	& \underline{1.75} & \textbf{1.25} \\
         \bottomrule
    \end{tabular}
\end{table}

\FloatBarrier
Overall, \emph{ZEUS+TabICL} performs competitively with the original ZEUS across all evaluated data collections and even surpasses it in rank on synthetic data. Nonetheless, its average ARI on real-world datasets is noticeably lower. Future work should investigate whether this is due to random seed variability, our incomplete understanding of the TabICL architecture, potential overfitting to the prior, or the need for a more expressive prior.

\subsection{Special-case scenarios}
Finally, we examine the behavior of our model in several extreme scenarios. To this end, we construct four synthetic two-dimensional datasets designed to highlight specific challenges:
\begin{itemize}
    \item \textbf{Cluster-Imbalance}: Three clusters with 1000, 300, and 50 samples, respectively.
    \item \textbf{Covariance-Scaled}: Three equally sized clusters (500 samples each) with covariance matrices scaled by factors of 1, 6, and 36.
    \item \textbf{Moons}: Generated using \emph{make\_moons} from \emph{scikit-learn} with 1000 samples and 0.1 noise.
    \item \textbf{Connection}: Two spherical Gaussians linked by a degenerate linear Gaussian.
\end{itemize}
We report a comparison of \our{} performance against $K$-Means and GMM in Table~\ref{tab:spec_cases}.

\begin{table}[!ht]
    \centering
    \caption{Comparison of clustering quality on synthetic special-case datasets, assessed by the ARI metric (higher scores indicate better quality).}
    \label{tab:spec_cases}
    \begin{tabular}{cccc}
        \toprule
        & KM & GMM & ZEUS \\
        \midrule
        Cluster-Imbalance & 50.12 & \underline{96.75} & \textbf{96.80} \\ 
        Covariance-scaled & 79.68 & \textbf{96.45} & \underline{94.72} \\
        Moons & 48.67 & \underline{49.23} & \textbf{98.80} \\ 
        Connection & 72.76 & \textbf{89.30} & \underline{83.16} \\ 
        \bottomrule
    \end{tabular}
\end{table}
\FloatBarrier

In summary, \our{} outperforms $K$-means across all cases and performs on par with GMM. Notably, on the Moons dataset, ZEUS is the only method that successfully separates the clusters. This suggests that these synthetic examples do not represent genuine failure cases for our approach. A more thorough analysis of datasets where ZEUS underperforms remains an important direction for future work.

\section{Licensing and Third-Party Assets.}
Our method builds upon the publicly available TabPFN codebase, which is released under the Apache 2.0 License. All real-world datasets used in the evaluation are sourced from OpenML.org, a platform hosting open datasets for machine learning research; all datasets used are publicly accessible and labeled as open data, although individual datasets may be subject to specific licenses (e.g., CC-BY or similar). We ensured that no proprietary or restricted datasets were used. Our released code is provided under the Apache 2.0 License and includes full instructions to reproduce the experiments.

\paragraph{Broader Impact.}
Although \our{} is designed as a general-purpose tool for clustering tabular data, its use in high-stakes domains like healthcare, finance, or criminal justice carries inherent risks. In such contexts, automated grouping of individuals - especially without labels or fairness constraints - can lead to biased or opaque outcomes. We therefore recommend that any deployment of \our{} in sensitive applications be accompanied by fairness assessment and expert review.

\end{document}